\documentclass{article}

\PassOptionsToPackage{numbers, compress}{natbib}
 \usepackage[preprint]{neurips_2026}

\usepackage{CJKutf8} 
\usepackage[utf8]{inputenc} 
\usepackage[T1]{fontenc}    
\usepackage{hyperref}       
\usepackage{url}            
\usepackage{booktabs}       
\usepackage{amsfonts}       
\usepackage{nicefrac}       
\usepackage{microtype}      
\usepackage{xcolor}         
\usepackage[table]{xcolor}
\usepackage{wrapfig}

\usepackage{bm}

\usepackage{subcaption} 
\usepackage{multirow}
\usepackage{graphicx}
\usepackage{amsmath}
\usepackage{cleveref}

\definecolor{deepiris}{HTML}{5D3FD3}

\newcommand{\greybg}[1]{{\cellcolor[HTML]{D0CECE}\textbf{#1}}}

\title{Rethinking Constraint Awareness for Efficient State Embedding of Neural Routing Solver}

\author{%
  Canhong Yu$^{1}$ \quad Changliang Zhou$^{2,3}$ \quad Rongsheng Chen$^{2,3,4}$ \quad Zhenkun Wang$^{2,3}$ \quad Yu Zhou$^{1}$ \\
  $^{1}$ College of Computer Science and Software Engineering, Shenzhen University, Shenzhen, China\\
  $^{2}$ School of Automation and Intelligent Manufacturing,\\ Southern University of Science and Technology, Shenzhen, China\\
  $^{3}$ Guangdong Provincial Key Laboratory of Fully Actuated System Control Theory and Technology,\\
  Southern University of Science and Technology, Shenzhen, China\\
  $^{4}$ Pengcheng Laboratory, Shenzhen, China\\
  \texttt{2410103054@mails.szu.edu.cn, \{zhoucl2022,chenrs2025\}@mail.sustech.edu.cn,}\\
  \texttt{wangzk3@sustech.edu.cn,}
  \texttt{zhouyu\_1022@126.com}
}

\begin{document}
\begin{CJK*}{UTF8}{gbsn} 

\maketitle

\begin{abstract}
Heavy-Encoder-Light-Decoder (HELD) neural routing solvers have emerged as a promising paradigm due to their broad applicability across multiple vehicle routing problems (VRPs). However, they typically struggle with VRP variants with complex constraints. To address this limitation, this paper systematically revisits existing neural solvers from the perspective of the generation mechanism for state embeddings (i.e., query vector prior to compatibility calculation) during decoding. We identify that current mechanisms restrict the observation space during attention computation, introducing a key bottleneck to achieving high-quality solutions. Through detailed empirical analysis, we demonstrate the necessity of preserving a global observation space. To overcome the constraint-agnostic drawback inherent to global observation spaces, we propose a simple yet powerful Constraint-Aware Residual Modulation (CARM) module. By adaptively modulating the context embedding with constraint-relevant variables, CARM effectively enhances constraint awareness, enabling the neural solver to fully leverage the global observation space and generate an efficient state embedding. Extensive experimental results across two single-task and five multi-task neural routing solvers confirm that the CARM module consistently boosts baseline performance. Notably, solvers equipped with our CARM achieve substantial improvements in scaling to large-scale instances and in generalizing to unseen VRP variants. These findings provide valuable insights for the architectural design of neural routing solvers.

\end{abstract}

\section{Introduction}
The vehicle routing problem (VRP) is an important class of combinatorial optimization problems (COPs) with broad applications in logistics, supply chain management, and related domains~\citep{tiwari2023optimization,sar2023systematic}. Owing to its NP-hardness, obtaining optimal solutions via exact solvers for VRP instances is often computationally prohibitive. While traditional leading heuristics~\citep{LKH3,HGS} have achieved near-optimal performance within acceptable computation time, they rely on handcrafted rules, whose design typically requires substantial domain expertise. Additionally, real-world applications often impose distinct constraints that give rise to diverse VRP variants, which render the manual design of tailored rules for each variant practically untenable.
 
Recently, neural combinatorial optimization (NCO) methods have emerged as a promising alternative for solving VRPs~\citep{bengio2021machine,ba2026survey}. These methods leverage neural networks to learn implicit heuristic rules directly from data, thereby reducing reliance on expert knowledge while maintaining competitive solution quality~\citep{li2024fastT2T,sun2023difusco,qiu2022dimes,fang2024invit,zhou2025L2R,li2025drhg,ye2023glop,ye2023deepaco,zheng2024udc}. Among them, constructive solvers have attracted significant attention due to their conceptual simplicity and adaptability to various VRP variants, which develop into two mainstream Transformer~\citep{vaswani2017attention}-based architectural paradigms: Heavy-Encoder-Light-Decoder (HELD)~\citep{kool2019attention,kwon2020pomo,zhou2024icam} and Light-Encoder-Heavy-Decoder (LEHD)~\citep{luo2023lehd,drakulic2023bq,luo2024Boosting}. HELD-based solvers typically employ a heavy encoder to generate one-shot static node embeddings and subsequently construct a complete solution based on a light decoder in an autoregressive manner. To date, the HELD architecture has driven the development of numerous single-task neural solvers~\citep{gao2023elg,jin2023pointerformer,kwon2021matnet,kim2022symnco} tailored to specific problems, alongside recently emerging multi-task neural solvers~\citep{berto2024routefinder,li2025cada,zhou2025urs} capable of handling diverse VRP variants. While widely adopted for their high computational efficiency and ease of integration with reinforcement learning (RL) algorithms, they struggle to achieve satisfactory performance on various VRP variants. A detailed review of related work on single-task and multi-task learning is provided in Appendix \ref{app:related_work}.

\begin{figure*}[t]
\begin{center}
\centerline{\includegraphics[width=\linewidth]{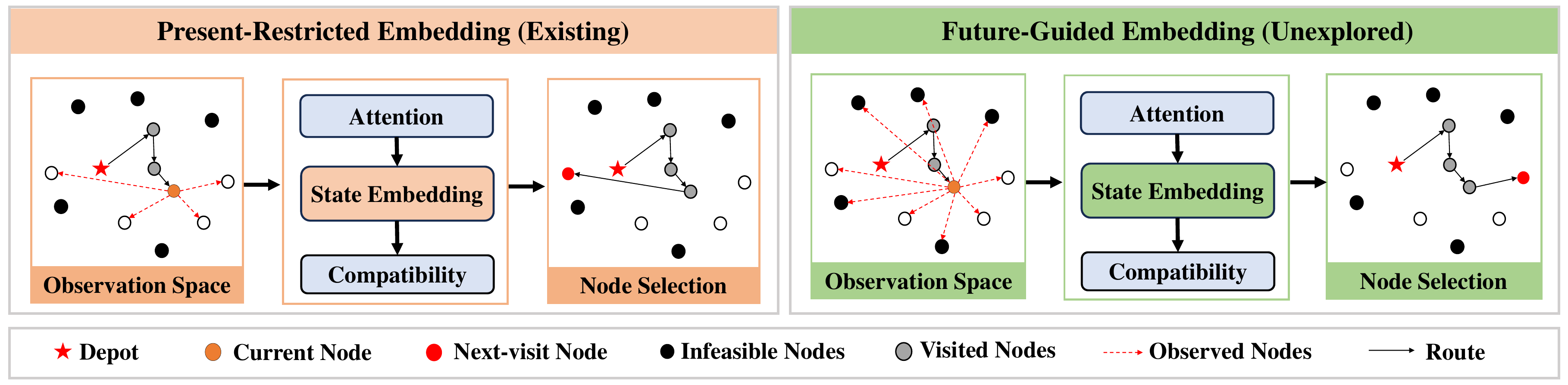}}
\caption{\textbf{Comparison of different state embedding generation mechanisms.} Left: Existing mechanism, which we refer to as present-restricted embedding (PRE), limits attention to currently feasible nodes, which lacks a global perspective and results in poor node selection. Right: The future-guided embedding (FGE) expands attention to all potential candidate nodes (i.e., unvisited nodes), which is grounded in a global observation space and delivers better node selection than PRE. }
\label{fig:PRE_FGE}
\end{center}
\vskip -0.4in
\end{figure*}

When solving VRP variants with diverse constraints, state embedding plays a pivotal role because it embeds constraint-relevant variables and directly dictates node selection during decoding. As illustrated in Figure \ref{fig:PRE_FGE}, current HELD-based solvers~\citep{kool2019attention,kwon2020pomo} typically construct a present-restricted embedding (PRE) by applying an attention mechanism~\citep{vaswani2017attention} between the current state and nodes that are currently feasible at each step. It then serves as the sole dynamically updated input to the subsequent compatibility calculation, which selects the next feasible node. The restricted embedding deprives the model of a necessary global perspective, resulting in poor node selection. Notably, by expanding attention to all potential candidate nodes required for the final solution (i.e., all unvisited nodes), the future-guided embedding (FGE) constructs a state embedding grounded in a global observation space, thereby delivering better node selection than traditional PRE. We provide a comprehensive empirical result and analysis in Section \ref{sec:diff_embedding}.

In this paper, we systematically revisit existing HELD-based neural solvers from the perspective of the generation mechanism for state embeddings. First, we identify that the PRE-induced restricted observation space introduces a critical bottleneck to achieving high-quality solutions. Second, our empirical analysis reveals that while preserving a global observation space enables neural solvers to achieve better results across various seen VRP variants than their vanilla counterparts, it hinders the ability to distinguish among constraints in the observation space, thereby resulting in unsatisfactory performance on unseen VRPs. Finally, motivated by these findings, we propose a simple yet powerful \textbf{\underline{C}}onstraint-\textbf{\underline{A}}ware \textbf{\underline{R}}esidual \textbf{\underline{M}}odulation (CARM) to improve the constraint awareness of neural solvers, enabling the neural solver to fully leverage the global observation space and generate an efficient state embedding. We evaluate our CARM across multiple single-task and multi-task neural solvers. Extensive empirical results show that the proposed CARM consistently boosts baseline performance across both seen and unseen VRPs. Our main contributions are summarized as follows:

\begin{itemize}
    \item  We systematically revisit the existing state embedding generation mechanism and identify that the prevalent PRE restricts the observation space during attention computation, introducing a key bottleneck to achieving high-quality solutions.
    \item  We demonstrate the necessity of preserving a global observation space and propose a simple yet powerful CARM module to enhance constraint awareness of neural solvers, thereby fully leveraging global observation space and generating an efficient state embedding. 
    \item Extensive experimental results across two single-task and five multi-task neural routing solvers confirm that the CARM module consistently boosts baseline performance. Notably, solvers equipped with our CARM achieve substantial improvements in scaling to large-scale instances and in generalizing to unseen VRP variants.
\end{itemize}

\section{Preliminaries}
\label{sec:preliminaries}
In this section, we first introduce the definition of VRP, then provide an overview of the solution generation process for HELD-based constructive neural solvers~\citep{kool2019attention,kwon2020pomo,zhou2024icam}.
\subsection{Vehicle Routing Problem}
A VRP instance can be defined as a graph $\mathcal{G}=(\mathcal{V}, \mathcal{E})$, where the node set $\mathcal{V}=\{v_i\}_{i=0}^n$ includes an optional depot indexed by $0$ and $n$ customers, and the edge set is $\mathcal{E}=\{e(v_i,v_j)\ |\ v_i,v_j\in \mathcal{V}, v_i \neq v_j\} $. A feasible solution is a finite sequence that satisfies all problem-specific constraints, denoted as $\bm{\pi} = (\pi_1, \pi_2, \ldots, \pi_T)$, where $T$ is the total number of construction steps. For example, in the Capacitated VRP (CVRP), it is feasible if each customer node is visited exactly once and the total demand in each sub-tour does not exceed vehicle capacity. Let $\Omega$ denote the feasible sequence set, our goal is to find the optimal sequence $\bm{\pi}^* = \arg\max_{\bm{\pi}\in \Omega} f(\bm{\pi}|\mathcal{G})$, where the objective function $f(\bm{\pi} | \mathcal{G})$ is typically defined as the negative total distance of $\bm{\pi}$.

\subsection{HELD-based Neural Routing Solver}
Let $\bm{\theta}=\{\bm{\theta}_{enc},\bm{\theta}_{dec}\}$ denote the learnable parameters of the encoder and decoder, respectively. Its key components are elaborated below. The detailed description is provided in Appendix \ref{app:held_model_Architecture}.

\paragraph{Model Architecture}
Given a VRP instance $\mathcal{G}$, the raw node features are first mapped to initial embeddings $H^{(0)}=\{\mathbf{h}_i^{(0)}\}_{i=0}^n \in \mathbb{R}^{(1+n)\times d}$ via a linear projection, where $d$ is the dimension of the embeddings. $H^{(0)}$ then processed by an $L$-layer attention-based encoder~\citep{kool2019attention} $\bm{\theta}_{enc}$ to produce advanced node embeddings $H^{(L)}=\{\mathbf{h}_i^{(L)}\}_{i=0}^n \in \mathbb{R}^{(1+n)\times d}$. $\bm{\theta}_{dec}$ comprises state-updating and compatibility computation modules. Among them, the state-updating module plays a pivotal role because it embeds constraint-relevant variables $C_t \in \mathbb{R}^{m}$ (e.g., the remaining load in CVRP with $m=1$) and directly dictates node selection. For the example of CVRP, it takes $H^{(L)}$ and context embedding $\mathbf{h}_{(C)}^{t}=[\mathbf{h}_{\pi_{t-1}}^{(L)}, C_t]\in \mathbb{R}^{d+m}$ as inputs and leverages an attention~\citep{kool2019attention} to generate an advanced state embedding $\hat{\mathbf{h}}_{(C)}^{t}\in \mathbb{R}^{d}$, where $\mathbf{h}_{\pi_{t-1}}^{(L)} \in \mathbb{R}^{d}$ is the node embedding of the last visited node. $\hat{\mathbf{h}}_{(C)}^{t}$ is then used to evaluate its compatibility with all nodes and select the next feasible node. $\bm{\theta}_{dec}$ builds a solution incrementally from scratch by sequentially appending the next feasible node to the current partial solution $\pi_{1:t-1}=(\pi_{1},\pi_{2},...,\pi_{t-1})$, until a complete solution is constructed. 

\paragraph{Masking Mechanism}
At each decoding step $t$, neural solvers employ a model-agnostic masking function $\mathcal{M}_{t}$. For the vanilla attention with $\mathcal{M}_{t}$~\citep{kool2019attention}, the calculation can be expressed as:
\begin{equation}
\mathrm{Attention}(Q,K,V) = \mathrm{Softmax}\left(\frac{QK^T}{\sqrt{d}} + \mathcal{M}_{t}\right)V,
\label{eq:attention_mask}
\end{equation}
where $Q$, $K$, and $V$ denote the query, key, and value matrices, respectively. To ensure the feasibility of the solutions, the probabilities of currently infeasible (i.e., constraint-violating) nodes are set to zero by assigning $-\infty$ bias during the compatibility computation. Beyond this mandatory step, PRE $\hat{\mathbf{h}}_{(C_p)}^t$ assigns $-\infty$ bias in $\mathcal{M}_{t}$ to visited and infeasible nodes, fully eliminating their effects on the decision. Conversely, FGE $\hat{\mathbf{h}}_{(C_f)}^t$ retains currently infeasible nodes and masks only visited nodes to preserve the Markov property, enabling attention to assess the potential importance of all candidates.

\section{Performance under Different State Embeddings}
\label{sec:diff_embedding}

Since multi-task learning handles diverse constraints simultaneously, its decoder is more sensitive to the state embedding than single-task models. To explore how different state embeddings affect model performance, we evaluate their impact on multi-task solvers across various VRP variants.

\subsection{Experimental Setup} 

\paragraph{Evaluated Solvers}
\label{sec:evaluated_solvers}
To show the broad applicability of our findings, we select three representative multi-task neural routing solvers: MTPOMO~\citep{liu2024mtpomo}, MVMoE-4E/L (abbreviated as MVMoE)~\citep{zhou2024mvmoe}, and ReLD-MTL~\citep{huang2025reld}. We evaluate their official pretrained models in the same problem setting and denote their default state embedding generation strategy as 'PRE'\footnote{For MTPOMO~\citep{liu2024mtpomo}, we use the implementation provided by MVMoE~\citep{zhou2024mvmoe}, ensuring consistency across all three models.}. To ensure a fair comparison, we retrain the three models with strictly identical training configurations and replace their default state embedding mechanisms with our FGE, denoted as 'FGE'. 

\paragraph{Problem \& Datasets}
\label{sec:benchmark_tasks}

In this experiment, we test 16 distinct VRP variants widely studied in existing research~\citep {liu2024mtpomo,zhou2024mvmoe,huang2025reld}. Each variant incorporates one or more of the following constraints: Capacity (C), Open Route (O), Backhaul (B), Duration Limit (L), and Time Windows (TW). We strictly adhere to the constraint settings established in~\cite{zhou2024mvmoe,huang2025reld} for fair evaluation (see Appendix \ref{app:setups_of_vrp_variants} for details). To align with current mainstream evaluation standards, we use VRP instances with 100 nodes, along with the datasets provided by \citep{zhou2024mvmoe}, with each containing 1,000 instances.

\paragraph{Metrics \& Inference}
\label{sec:metrics_and_inference}
We report the optimality gap (Gap) across all methods, which quantifies the discrepancy between the solutions generated by each method and the near-optimal solutions obtained with classical solvers. We report the best result of $n$-trajectory with $\times 8$ instance augmentation~\citep{kwon2020pomo}.
\begin{figure}[t] 
    \centering

    \begin{subfigure}{0.24\textwidth}
        \centering
        \includegraphics[width=\textwidth]{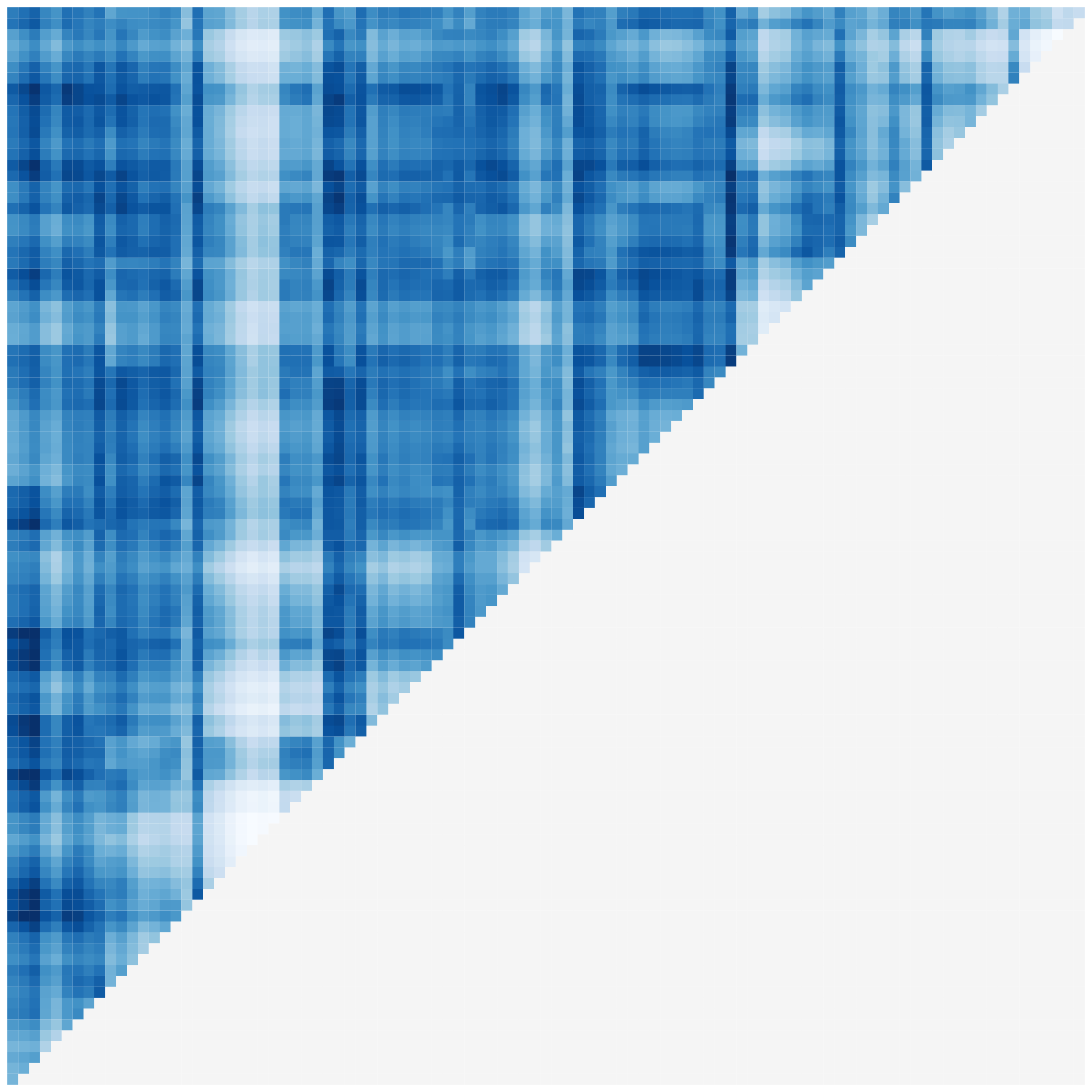}
        \caption{\centering CVRP under PRE}
        \label{subfig:heatmap_cvrp_full}
    \end{subfigure}\hfill
    \begin{subfigure}{0.24\textwidth}
        \centering
        \includegraphics[width=\textwidth]{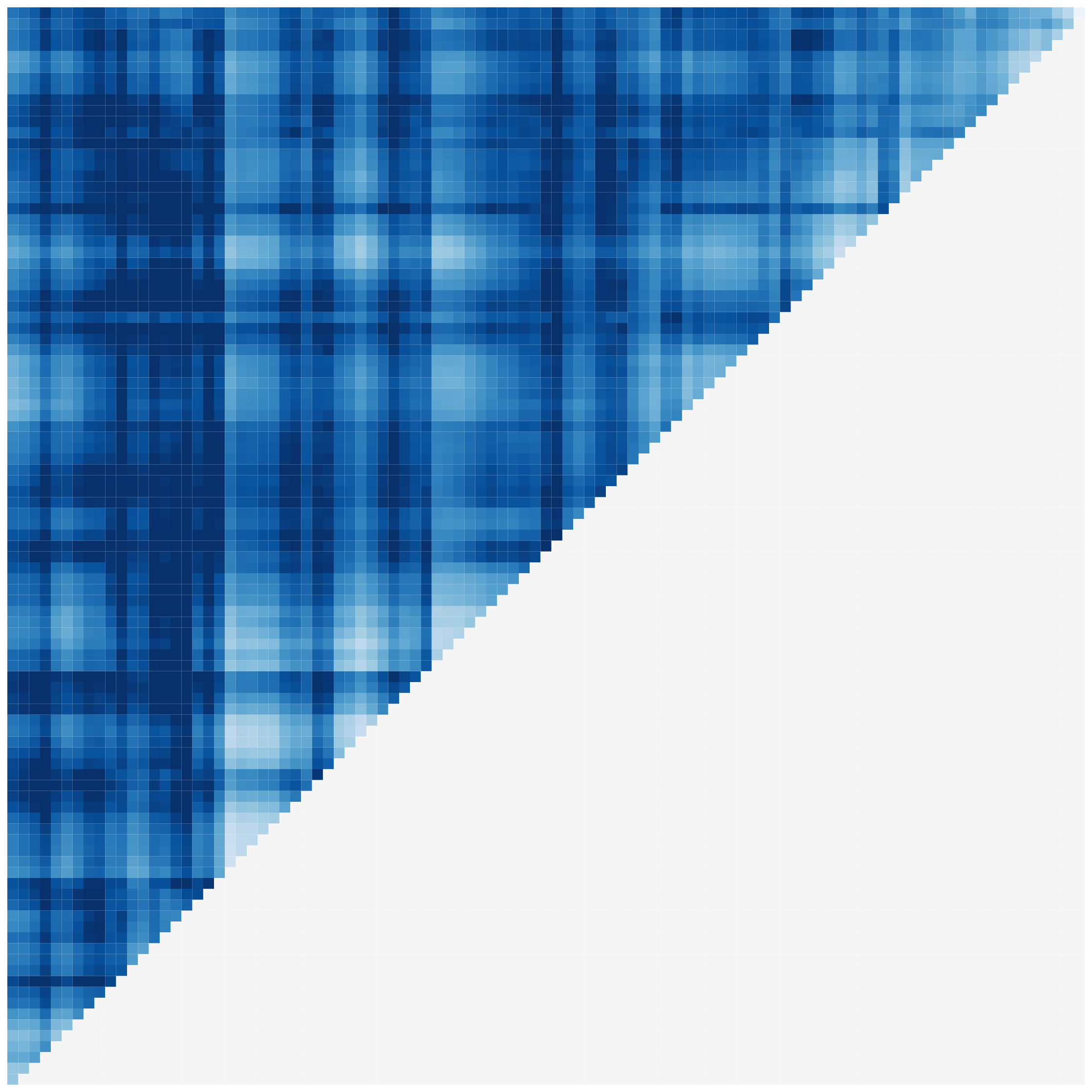}
        \caption{\centering CVRP under FGE}
        \label{subfig:heatmap_cvrp_visited}
    \end{subfigure}\hfill
    \begin{subfigure}{0.24\textwidth}
        \centering
        \includegraphics[width=\textwidth]{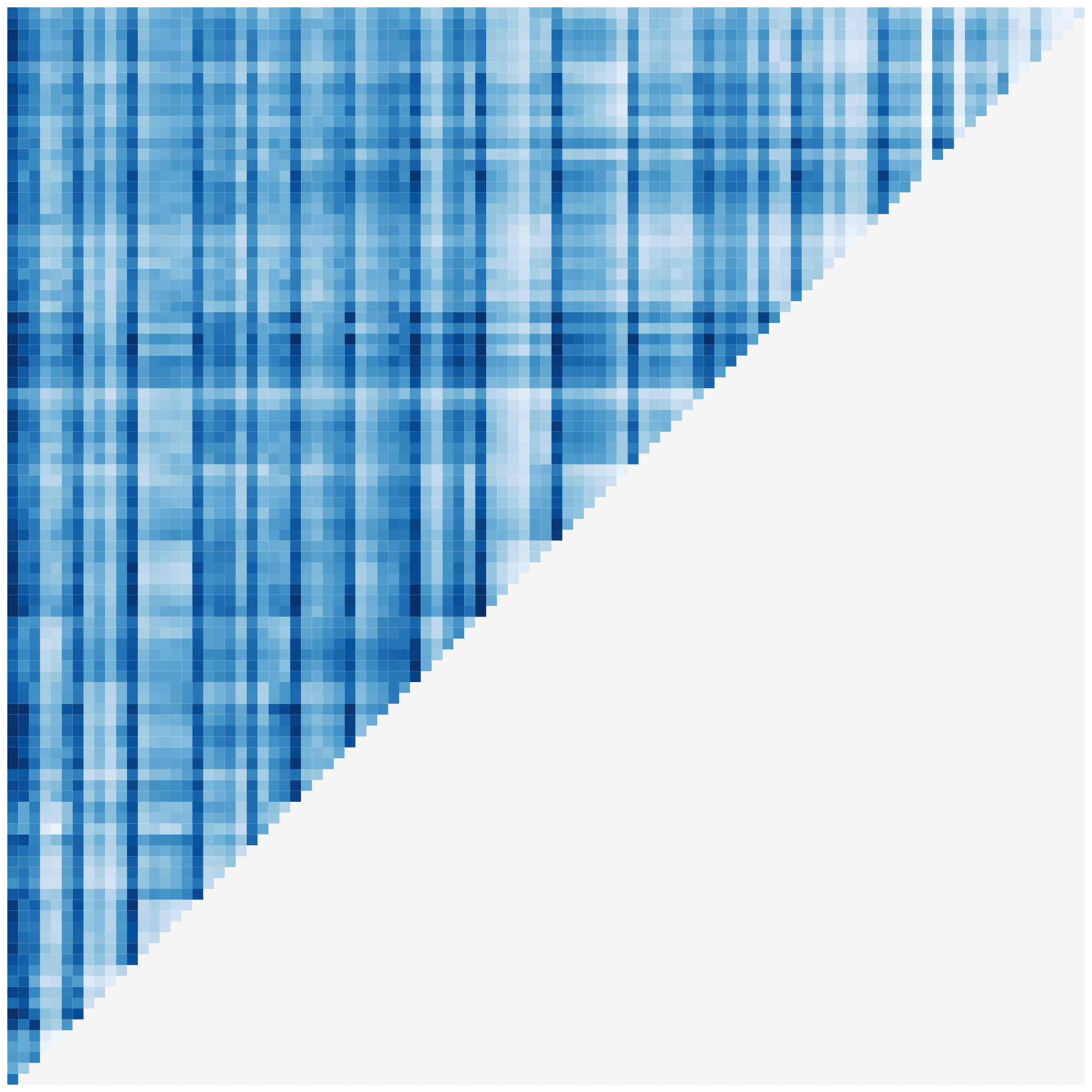}
        \caption{\centering CVRPTW under PRE}
        \label{subfig:heatmap_cvrptw_full}
    \end{subfigure}\hfill
    \begin{subfigure}{0.24\textwidth}
        \centering
        \includegraphics[width=\textwidth]{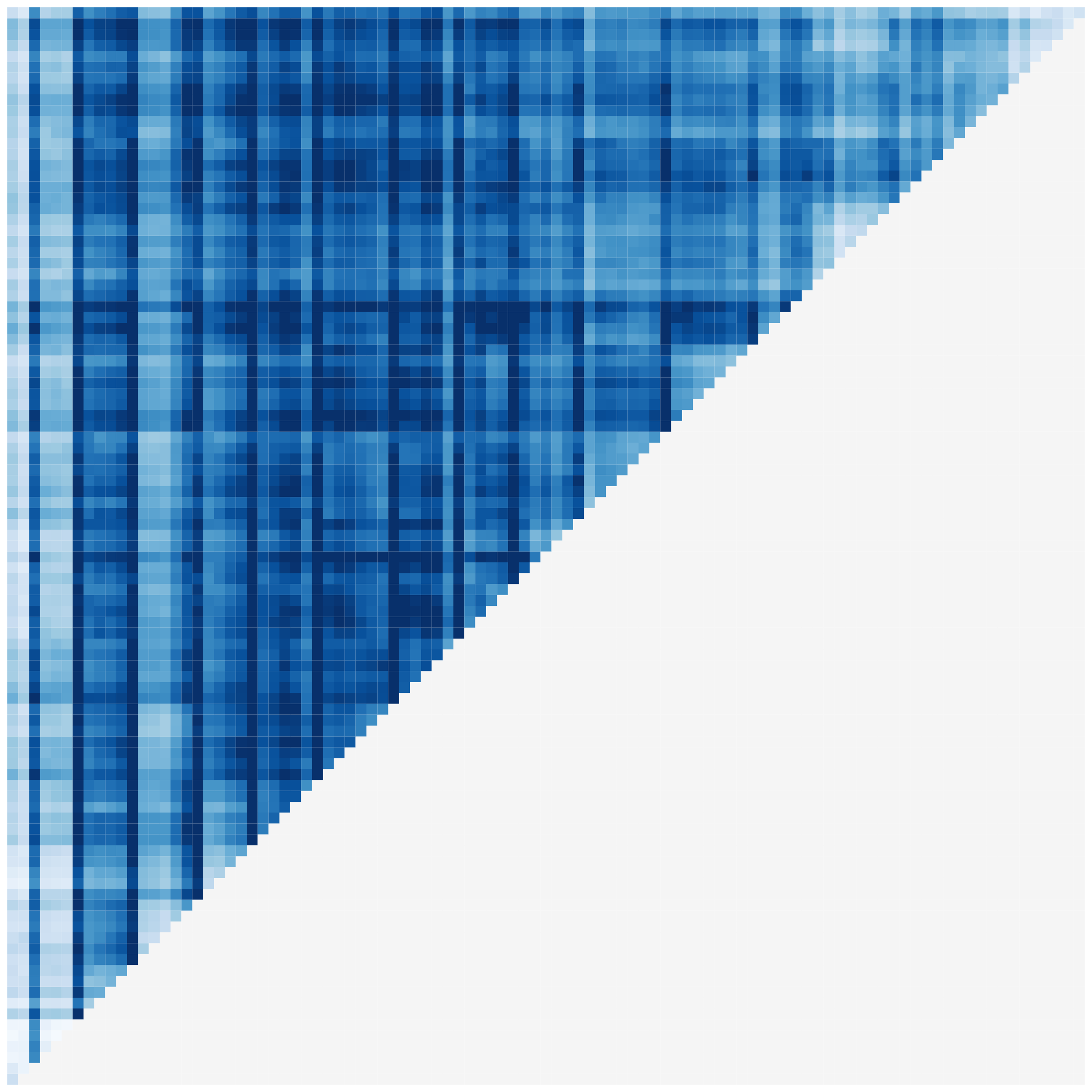}
        \caption{\centering CVRPTW under FGE}
        \label{subfig:heatmap_cvrptw_visited}
    \end{subfigure}
   
    \caption{Comparison of Gaussian similarities between state embedding $\hat{\mathbf{h}}_{(C)}^{t}$ and node embeddings $H^{(L)}$ generated by PRE and FGE at each decoding step, evaluated using ReLD-MTL. A darker shade indicates higher similarity. Note that we mask the similarities for visited nodes. For detailed experimental setup and more results, please refer to Appendix \ref{app:heatmaps}.}
    \label{fig:heatmap_cvrp_cvrptw}
\end{figure}

\subsection{Results and Analyses}

\begin{wraptable}[11]{r}{0.6\textwidth}
\vspace{-18pt}
\begin{center}

\caption{Optimality gap comparison of different state embedding generation strategies on 6 seen VRP variants.}
\resizebox{\linewidth}{!}{
    \begin{tabular}{l|cc|cc|cc}
    \toprule[0.5mm]
    \multirow{2}{*}{Problem} & \multicolumn{2}{c|}{MTPOMO} & \multicolumn{2}{c|}{MVMoE} & \multicolumn{2}{c}{ReLD-MTL} \\
     & PRE & FGE & PRE & FGE & PRE & FGE \\
    \midrule
    CVRP & 1.85\% & \cellcolor[HTML]{D0CECE}\textbf{1.83\%} & 1.73\% & \cellcolor[HTML]{D0CECE}\textbf{1.65\%} & 1.42\% & \cellcolor[HTML]{D0CECE}\textbf{1.37\%} \\
    OCVRP & \cellcolor[HTML]{D0CECE}\textbf{3.46\%} & 3.51\% & 3.21\% & \cellcolor[HTML]{D0CECE}\textbf{3.09\%} & 2.32\% & \cellcolor[HTML]{D0CECE}\textbf{2.26\%} \\
    CVRPB & 1.67\% & \cellcolor[HTML]{D0CECE}\textbf{1.48\%} & 1.37\% & \cellcolor[HTML]{D0CECE}\textbf{1.19\%} & 0.90\% & \cellcolor[HTML]{D0CECE}\textbf{0.68\%} \\
    CVRPTW & 5.31\% & \cellcolor[HTML]{D0CECE}\textbf{4.98\%} & 4.93\% & \cellcolor[HTML]{D0CECE}\textbf{4.59\%} & 4.56\% & \cellcolor[HTML]{D0CECE}\textbf{4.16\%} \\
    CVRPL & 0.48\% & \cellcolor[HTML]{D0CECE}\textbf{0.46\%} & 0.32\% & \cellcolor[HTML]{D0CECE}\textbf{0.26\%} & 0.02\% & \cellcolor[HTML]{D0CECE}\textbf{-0.02\%} \\
    OCVRPTW & 4.41\% & \cellcolor[HTML]{D0CECE}\textbf{3.93\%} & 3.94\% & \cellcolor[HTML]{D0CECE}\textbf{3.39\%} & 3.10\% & \cellcolor[HTML]{D0CECE}\textbf{2.62\%} \\
    \midrule
    Avg. Gap & 2.86\% & \cellcolor[HTML]{D0CECE}\textbf{2.70\%} & 2.58\% & \cellcolor[HTML]{D0CECE}\textbf{2.37\%} & 2.05\% & \cellcolor[HTML]{D0CECE}\textbf{1.85\%} \\
    Best Solution & 1/6 & \cellcolor[HTML]{D0CECE}\textbf{5/6} & 0/6 & \cellcolor[HTML]{D0CECE}\textbf{6/6} & 0/6 & \cellcolor[HTML]{D0CECE}\textbf{6/6} \\
    \bottomrule[0.5mm]
    \end{tabular}%
  }
  \label{tab:mtpomo_mvmoe_reld_InD}%
\end{center}
\end{wraptable}
\paragraph{Performance on Seen VRPs} 
As shown in Table \ref{tab:mtpomo_mvmoe_reld_InD}, replacing the existing PRE with FGE greatly improves the overall performance of all solvers across 6 seen VRPs. Notably, MVMoE and ReLD-MTL under FGE perform better across all variants than PRE. As illustrated in Figure \ref{fig:heatmap_cvrp_cvrptw}, FGE $\hat{\mathbf{h}}_{(C_f)}^{t}$ exhibits a stronger correlation with $H^{(L)}$, which shows that FGE effectively captures global state during decoding, thereby enabling the model to perform better than PRE (see Appendix \ref{app:heatmaps} for detailed discussion). From the perspective of the observation space, the two strategies differ fundamentally:

\begin{itemize}

    \item \textbf{Step-wise Myopia of PRE:} Existing PRE inherently relies on a local observation space, which obscures the global structural features of the instances. The model is forced to evaluate the immediate utility of each feasible node from a restricted perspective, which provides myopic node selection and results in unsatisfactory performance, especially when facing complex VRP variants (e.g., OCVRPTW). 
    \item \textbf{Trajectory-wise Foresight of FGE:}  In contrast, FGE enables the model to generate the state embedding under a global observation space. By assessing the potential importance of all potential candidate nodes, the attention mechanism can produce an efficient state embedding from a global perspective. Given that the objective of VRPs is to optimize the total tour rather than individual steps, FGE offers a better trade-off between long-term rewards and sequential decision-making, thereby delivering better overall performance than PRE.

\end{itemize}

\begin{wraptable}[15]{r}{0.6\textwidth}
\vspace{-18pt}
\begin{center}

\caption{Zero-shot generalization of different state embedding generation strategies on 10 unseen VRP variants.}
\resizebox{\linewidth}{!}{
    \begin{tabular}{l|cc|cc|cc}
    \toprule[0.5mm]
    \multirow{2}{*}{Problem} & \multicolumn{2}{c|}{MTPOMO} & \multicolumn{2}{c|}{MVMoE} & \multicolumn{2}{c}{ReLD-MTL} \\
    
    & PRE & FGE & PRE & FGE & PRE & FGE \\
    \midrule
    OCVRPB    & \greybg{7.34\%} & 8.06\% & 7.24\% & \greybg{7.00\%} & 5.36\% & \greybg{5.31\%} \\
    CVRPBL    & 1.79\% & \greybg{1.44\%} & 1.47\% & \greybg{1.20\%} & 1.01\% & \greybg{0.69\%} \\
    CVRPLTW   & 1.92\% & \greybg{1.57\%} & 1.55\% & \greybg{1.23\%} & 1.17\% & \greybg{0.81\%} \\
    OCVRPBTW  & \greybg{10.45\%}& 10.96\%& \greybg{10.19\%}& 10.39\% & \greybg{9.29\%} & 9.50\% \\
    CVRPBLTW  & \greybg{7.75\%} & 8.17\% & \greybg{7.47\%} & 7.76\% & \greybg{6.94\%} & 7.04\% \\
    OCVRPL    & \greybg{3.44\%} & 3.99\% & 3.24\% & \greybg{3.13\%} & 2.31\% & \greybg{2.28\%} \\
    OCVRPBLTW & \greybg{10.50\%}& 11.03\%& \greybg{10.26\%}& 10.45\% & \greybg{9.22\%} & 9.50\% \\
    CVRPBTW   & \greybg{7.41\%} & 8.24\% & \greybg{7.19\%} & 7.35\% & 6.74\% & \greybg{6.71\%} \\
    OCVRPBL   & \greybg{7.34\%} & 8.11\% & 7.30\% & \greybg{7.08\%} & 5.41\% & \greybg{5.31\%} \\
    OCVRPLTW  & 4.37\% & \greybg{3.97\%} & 3.97\% & \greybg{3.48\%} & 3.16\% & \greybg{2.63\%} \\
    \midrule
    Avg. Gap  & \greybg{6.23\%} & 6.55\% & 5.99\% & \greybg{5.91\%} & 5.06\% & \greybg{4.98\%} \\
    Best Solution & \greybg{7/10}   & 3/10   & 4/10   & \greybg{6/10}   & 3/10   & \greybg{7/10} \\
    \bottomrule[0.5mm]
    \end{tabular}%
  }
  \label{tab:mtpomo_mvmoe_reld_OoD}%
\end{center}
\end{wraptable}

\paragraph{Performance on Unseen VRPs}

As shown in Table \ref{tab:mtpomo_mvmoe_reld_OoD}, the advantages of FGE do not extend to the zero-shot generalization on unseen variants. Fundamentally, the observation space is controlled by $\mathcal{M}_t$ (see Equation \ref{eq:attention_mask}). While FGE retains richer instance information, it makes $\mathcal{M}_t$ constraint-agnostic, i.e., the model confronts identical observation spaces even when the constraints change. For example, regardless of the vehicle's remaining load, the candidates visible to the attention remain the same. Because attention computation fails to provide differential information reflecting current constraints, the burden of distinguishing varying constraints falls on the context embedding $\mathbf{h}_{(C)}^{t}$ that integrates constraint variables $C_t$.

While this shortcoming can be mitigated through extensive training on seen problems, it is apparent in zero-shot generalization settings. If constraint variables are simply concatenated to form $\mathbf{h}_{(C)}^{t}$, the resulting $\hat{\mathbf{h}}_{(C_f)}^{t}$ struggles to distinguish diverse constraints under this identical space, which leads to poor performance of the FGE-based model on unseen problems, as shown in the results of MTPOMO~\citep{liu2024mtpomo}. Recent solvers like MVMoE~\citep{zhou2024mvmoe} and ReLD-MTL~\citep{huang2025reld} aim to obtain more powerful $\hat{\mathbf{h}}_{(C_f)}^{t}$ by introducing gating mechanisms and constraint-based identity mapping functions after computing attention, respectively. While these methods improve performance in unseen problems under FGE strategies, they still struggle to outperform PRE as clearly as they do on seen problems.

These results demonstrate that simply adopting FGE results in performance degradation when facing unseen VRP variants. To better leverage the advantages of FGE and achieve superior performance across diverse VRP variants, a model must effectively distinguish changing constraint states under an unchanged observation space. This highlights the critical need for $\mathbf{h}_{(C)}^{t}$ with powerful constraint awareness and directly motivates our proposed CARM module.

\section{Constraint-Aware Residual Modulation}
\label{sec:methodology}

\begin{figure*}[t]
\begin{center}
\centerline{\includegraphics[width=\linewidth]{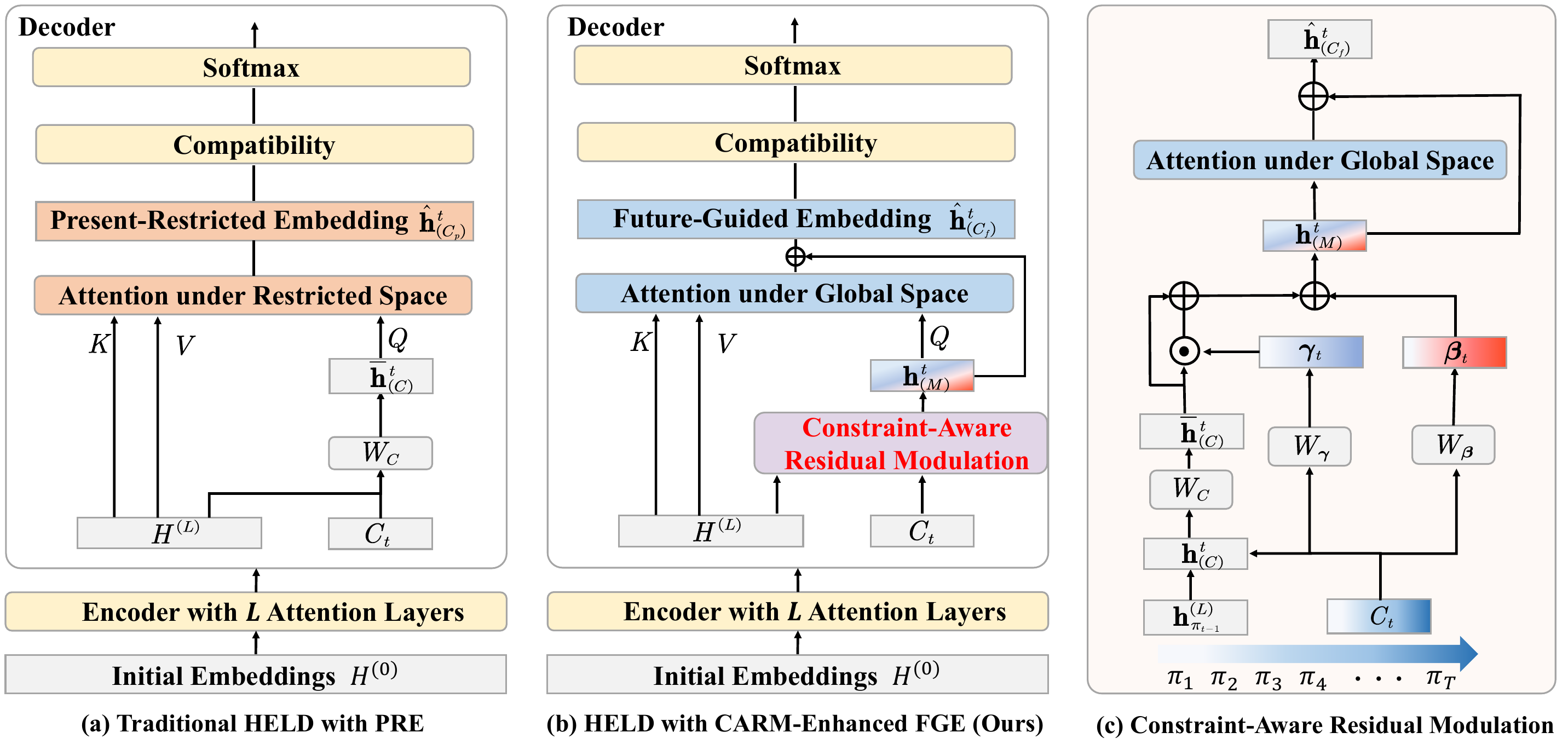}}
\caption{ \textbf{Comparison of different HELD architectures. }(a) \textbf{Traditional HELD with PRE:} Employs PRE with restricted observation space and lacks robust constraint awareness. (b) \textbf{Our HELD with CARM-Enhanced FGE:} Integrates FGE for global observation space and CARM for improved constraint awareness. (c) \textbf{CARM Calculation:} An efficient constraint-conditioned modulation mechanism by modulating context embedding $\mathbf{h}_{(C)}^{t}$ based on constraint-relevant variables. }
\label{fig:CARM}
\end{center}
\vskip -0.2in
\end{figure*}

In this section, we propose a novel HELD architecture integrated with a CARM-enhanced FGE to improve the performance of neural routing solvers, especially when tackling complex VRPs. As illustrated in Figure~\ref{fig:CARM}, by dynamically modulating $\mathbf{h}_{(C)}^{t}$ based on the varying $C_{t}$, CARM substantially improves the constraint awareness of $\mathbf{h}_{(C)}^{t}$, thereby delivering an efficient state embedding $\hat{\mathbf{h}}_{(C_f)}^t$. The implementation details are elaborated below.

\subsection{Constraint-Conditioned Modulation}
\label{sec:carm}

Inspired by the Feature-wise Linear Modulation (FiLM) mechanism\footnote{For detailed description of FiLM, please see Appendix \ref{app:film}.}~\citep{perez2018film}, which is widely adopted in multi-modal large language models (LLMs)~\citep{llm_film1,llm_film2,llm_film3}, we introduce a constraint-conditioned modulation strategy. This strategy enables the neural routing solver to dynamically adjust its context embedding $\mathbf{h}_{(C)}^{t}$ in response to varying constraint information $C_t$. As shown in Figure \ref{fig:CARM}(c), $C_t$ is used to generate learnable scaling and shifting parameters, which are further employed to adaptively modulate the concatenated context embedding $\mathbf{h}_{(C)}^{t}=[\mathbf{h}_{\pi_{t-1}}^{(L)}, C_t]$. This constraint-conditioned affine transformation ensures that dynamic constraints explicitly control the state representation. The modulation process is formulated as follows:
\begin{equation}
    \bar{\mathbf{h}}_{(C)}^t = \mathbf{h}_{(C)}^{t}W_C , \quad \gamma_t =  C_tW_{\gamma}, \quad \beta_t = C_tW_{\beta},
\end{equation}
\begin{equation}
   \mathbf{h}_{(M)}^t = \bar{\mathbf{h}}_{(C)}^t \odot (1 + \gamma_t) + \beta_t ,
\end{equation}
where $W_C \in \mathbb{R}^{(d+m) \times d  }$, $W_{\gamma}\in \mathbb{R}^{m \times d }$, $W_{\beta} \in \mathbb{R}^{m \times d }$ are learnable weight matrices, $\gamma_t$ and $\beta_t$ represent the scaling and shifting vectors derived from $C_t$, respectively. $\odot$ indicates the element-wise product. This constraint-conditioned modulation allows the resulting modulated context embedding $\mathbf{h}_{(M)}^t$ to effectively distinguish varying constraint states under the identical observation spaces induced by FGE. Consequently, it achieves a more powerful constraint awareness than the initial context embedding $\mathbf{h}_{(C)}^t$. Then $\mathbf{h}_{(M)}^t$ is fed into an attention mechanism operating under the global observation space and obtains an efficient FGE $\hat{\mathbf{h}}_{(C_f)}^t$.

\subsection{Efficient State Embedding}

For the light decoder, the encoder-generated node embeddings $H^{(L)}$ and the modulated context embedding $\mathbf{h}_{(M)}^t$ are fed into a state-updating module to generate an advanced state embedding. Without loss of generality, taking the well-known POMO~\citep{kwon2020pomo} as an example, this module includes a standard multi-head attention (MHA) mechanism~\citep{vaswani2017attention}. This procedure is formulated as follows: 
\begin{equation}
    Q=\mathbf{h}_{(M)}^tW_Q, \quad
    K=H^{(L)}W_K,\quad
    V=H^{(L)}W_V,
\end{equation}
\begin{equation}
    \mathbf{h}_{(A)}^{t} =  \mathrm{Attention}(Q,K,V,\mathcal{M}_t),
\end{equation}
\begin{equation}
    \hat{\mathbf{h}}_{(C_f)}^t = \mathbf{h}_{(M)}^t +  \mathbf{h}_{(A)}^{t},
\end{equation}
where $W_Q$,$W_K$,$W_V \in \mathbb{R}^{d \times d }$ denote learnable matrices. We explicitly mask visited nodes via $\mathcal{M}_t$ to satisfy the Markov property, while retaining all potential candidate nodes to preserve the global observation space. The attention output $\mathbf{h}_{(A)}^{t}$ is integrated with $\mathbf{h}_{(M)}^{t}$ via a residual connection operation~\citep{he2016skip} to produce an efficient FGE-based state embedding $\hat{\mathbf{h}}_{(C_f)}^{t}$. This operation effectively preserves the previously established constraint-aware context embedding while integrating the global context captured by the attention. Finally, $\hat{\mathbf{h}}_{(C_f)}^t$ serves as the sole dynamically updated representation in the subsequent compatibility calculation to compute the probability $p_{\bm{\theta}}(\pi_t=i \mid \pi_{1:t-1}, \mathcal{G}, C_{t})$ of selecting each feasible node $i$, see Appendix \ref{append:model_decoder} for more details.

\section{Experiments}
\label{sec:experiments}
In this section, we comprehensively evaluate the proposed CARM module by integrating it into multiple representative single-task and multi-task neural routing solvers and assessing them across various VRP variants with diverse constraints. Furthermore, we conduct ablation studies to validate the individual contributions and synergistic effects of our core components. All experiments are conducted on a single NVIDIA GeForce RTX 4090 GPU (24GB memory).

\subsection{Experimental Setup}
\label{sec:experimental setup}
\paragraph{Evaluated Solvers}

To evaluate the broad applicability of the proposed CARM, we incorporate it into seven representative neural solvers: (1) \textbf{Single-task Solvers:} POMO~\citep{kwon2020pomo} and ReLD-STL~\citep{huang2025reld}, which are both widely adopted backbones; (2) \textbf{Multi-task Solvers:} Three high-performance solvers under problem settings established by \citep{zhou2024mvmoe}, which are MTPOMO~\citep{liu2024mtpomo}, MVMoE-4E/L (abbreviated as MVMoE)~\citep{zhou2024mvmoe} and ReLD-MTL~\citep{huang2025reld}, as well as two leading models under the RouteFinder~\citep{berto2024routefinder} protocol, including RouteFinder-TE (abbreviated as RF)~\citep{berto2024routefinder} and CaDA~\citep{li2025cada}.

\paragraph{Problem \& Datasets}

For single-task evaluation, we test two neural solvers on six VRP variants (CVRP, OCVRP, CVRPL, CVRPB, CVRPTW, OCVRPTW). We also evaluate their scalability at larger scales ($N \in \{200, 500, 1000\}$ and capacity is fixed at $50$) using datasets from \citep{zheng2025mtl_kd}. For multi-task evaluation, we assess five solvers across 16 VRP variants used in Section~\ref{sec:benchmark_tasks}. Given that the 16 variants are seen tasks for both RF~\citep{berto2024routefinder} and CaDA~\citep{li2025cada}, we further evaluate their zero-shot generalization on eight unseen variants featuring Mixed Backhaul (MB) constraints. To ensure fairness given inconsistent constraint definitions (detailed in Appendix~\ref{app:setups_of_vrp_variants}), we strictly follow MVMoE~\citep{zhou2024mvmoe} and RF~\citep{berto2024routefinder} protocols using their $N = 100$ test datasets. In our experiment, each test dataset contains 1,000 instances. We provide detailed Oracle solver settings for the used datasets in Appendix~\ref{app:datasets}. For all evaluated models and datasets, we provide the corresponding official links in \ref{append:licenses}.

\paragraph{Model \& Training Settings}
The CARM-enhanced variants of all solvers (e.g., MTPOMO-CARM) are obtained by replacing the PRE with FGE and integrating our CARM module (architectural details in Appendix~\ref{app:carm}). During training, we retain all original hyperparameters (see Appendix \ref{app:training_settings}). For multi-task training, MTPOMO~\citep{liu2024mtpomo}, MVMoE~\citep{zhou2024mvmoe}, and ReLD-MTL~\citep{huang2025reld} are trained on six VRPs used in our single-task training, whereas RF~\citep{berto2024routefinder} and CaDA~\citep{li2025cada} use all 16 variants.

\paragraph{Metrics \& Inference}
Consistent with Section \ref{sec:metrics_and_inference}, we report the optimality gap (Gap) based on the best result from $n$ trajectories using $\times 8$ instance augmentation~\citep{kwon2020pomo}. We omit reporting inference times, as the additional computational overhead introduced by integrating CARM is negligible compared to that of the original solvers. For comparable results and further analysis about inference time, please see Appendix \ref{app:inference_time_analysis}.

\begin{table}[t]
  \centering
  \caption{Comparison of POMO and ReLD-STL with and without the CARM on various VRPs.}
  \resizebox{\linewidth}{!}{
    \begin{tabular}{l|cccccc}
    \toprule[0.5mm]
    Method & CVRP  & OCVRP  & CVRPB  & CVRPTW & CVRPL  & OCVRPTW \\
    \midrule
    POMO-PRE  & 1.49\% & 2.19\% & 1.00\% & 4.31\% & 0.09\% & 2.47\% \\
    POMO-CARM & \greybg{1.32\%} & \greybg{1.99\%} & \greybg{0.58\%} & \greybg{3.81\%} & \greybg{-0.04\%} & \greybg{1.79\%} \\
    \midrule
    \midrule
    ReLD-STL-PRE & 1.24\% & 1.56\% & 0.37\% & 3.90\% & -0.18\% & 2.22\% \\
    ReLD-STL-CARM & \greybg{1.10\%} & \greybg{1.41\%} & \greybg{0.19\%} & \greybg{3.52\%} & \greybg{-0.27\%} & \greybg{1.50\%} \\
    \bottomrule[0.5mm]
    \end{tabular}%
    }
  \label{tab:single_task_result}%
\end{table}%

\begin{table}[t]
  \centering
  \caption{Scalability on larger-scale instances of different VRP variants. All models are trained with single-task learning, and the training scale is fixed to $100$.}
  \resizebox{\linewidth}{!}{
    \begin{tabular}{l|ccc|ccc}
    \toprule[0.5mm]
      Method    & CVRP200 & CVRP500 & CVRP1000 & CVRPL200 & CVRPL500 & CVRPL1000 \\
    \midrule
    POMO-PRE  & 4.04\% & 10.88\% & 21.27\% & 4.77\% & 13.94\% & 34.03\% \\
    POMO-CARM & \greybg{3.98\%} & \greybg{8.39\%} & \greybg{10.10\%} & \greybg{3.33\%} & \greybg{8.21\%} & \greybg{11.52\%} \\
    \midrule
    ReLD-PRE  & 3.44\% & 7.65\% & 9.81\% & 2.93\% & 6.43\% & 8.40\% \\
    ReLD-CARM & \greybg{3.27\%} & \greybg{7.25\%} & \greybg{9.12\%} & \greybg{2.83\%} & \greybg{6.36\%} & \greybg{8.20\%} \\
    \midrule
    \midrule
          & CVRPTW200 & CVRPTW500 & CVRPTW1000 & OCVRP200 & OCVRP500 & OCVRP1000 \\
    \midrule
    POMO-PRE  & 9.75\% & 29.74\% & 69.64\% & 10.56\% & 22.86\% & 33.65\% \\
    POMO-CARM & \greybg{7.41\%} & \greybg{16.13\%} & \greybg{22.07\%} & \greybg{9.00\%} & \greybg{20.30\%} & \greybg{26.26\%} \\
    \midrule
    ReLD-PRE  & 7.29\% & 14.29\% & 19.00\% & 6.96\% & 14.03\% & 17.47\% \\
    ReLD-CARM & \greybg{6.42\%} & \greybg{12.93\%} & \greybg{16.40\%} & \greybg{6.46\%} & \greybg{12.84\%} & \greybg{16.49\%} \\
    \midrule
    \midrule
          & OCVRPTW200 & OCVRPTW500 & OCVRPTW1000 & CVRPB200 & CVRPB500 & CVRPB1000 \\
    \midrule
    POMO-PRE  & 10.62\% & 22.95\% & 36.63\% & 0.39\% & 4.73\% & 11.38\% \\
    POMO-CARM & \greybg{8.86\%} & \greybg{19.41\%} & \greybg{28.71\%} & \greybg{-0.99\%} & \greybg{0.07\%} & \greybg{2.69\%} \\
    \midrule
    ReLD-PRE & 8.45\% & 17.78\% & 24.29\% & -1.29\% & 0.74\% & 6.62\% \\
    ReLD-CARM & \greybg{7.22\%} & \greybg{15.42\%} & \greybg{20.29\%} & \greybg{-1.90\%} & \greybg{-1.61\%} & \greybg{1.17\%} \\
    \bottomrule[0.5mm]
    \end{tabular}%
  }
  \label{tab:stl_large_scale}%
\end{table}%

\subsection{Results with Single-Task Learning}
\paragraph{Performance on Training Scale}

As shown in Table \ref{tab:single_task_result}, CARM consistently improves the performance of two single-task solvers across all evaluated VRP variants. This improvement suggests that the module effectively enhances constraint awareness in neural solvers, thereby providing superior node selection during decoding. Notably, this performance gain is particularly pronounced in variants featuring complex constraint combinations, such as OCVRPTW. These results validate CARM's broad applicability in single-task learning, highlighting its model-agnostic nature.

\paragraph{Scalability to Larger-scale VRPs}

\begin{wraptable}[18]{r}{0.55\textwidth}
\vspace{-18pt}
\begin{center}
\caption{Comparison of solvers with and without CARM across 16 VRPs (6 seen and 10 unseen variants). VRPs marked with an asterisk ($^*$) are seen tasks.}
\resizebox{\linewidth}{!}{
    \begin{tabular}{l|cc|cc|cc}
    \toprule[0.5mm]
    \multirow{2}[2]{*}{Problem} & \multicolumn{2}{c|}{MTPOMO} & \multicolumn{2}{c|}{MVMoE} & \multicolumn{2}{c}{ReLD-MTL} \\
          & PRE   & CARM  & PRE   & CARM  & PRE   & CARM \\
    \midrule
    CVRP$^*$  & 1.85\% & \greybg{1.59\%} & 1.73\% & \greybg{1.54\%} & 1.42\% & \greybg{1.34\%} \\
    OCVRP$^*$ & 3.46\% & \greybg{2.87\%} & 3.21\% & \greybg{2.77\%} & 2.32\% & \greybg{2.11\%} \\
    CVRPB$^*$ & 1.67\% & \greybg{1.11\%} & 1.37\% & \greybg{1.00\%} & 0.90\% & \greybg{0.67\%} \\
    CVRPTW$^*$ & 5.31\% & \greybg{4.54\%} & 4.93\% & \greybg{4.39\%} & 4.56\% & \greybg{4.09\%} \\
    CVRPL$^*$ & 0.48\% & \greybg{0.20\%} & 0.32\% & \greybg{0.16\%} & 0.02\% & \greybg{-0.04\%} \\
    OCVRPTW$^*$ & 4.41\% & \greybg{3.14\%} & 3.94\% & \greybg{2.88\%} & 3.10\% & \greybg{2.49\%} \\
    \midrule
    OCVRPB & 7.34\% & \greybg{6.25\%} & 7.24\% & \greybg{6.19\%} & 5.36\% & \greybg{5.09\%} \\
    CVRPBL & 1.79\% & \greybg{1.11\%} & 1.47\% & \greybg{0.98\%} & 1.01\% & \greybg{0.65\%} \\
    CVRPLTW & 1.92\% & \greybg{1.20\%} & 1.55\% & \greybg{1.00\%} & 1.17\% & \greybg{0.71\%} \\
    OCVRPBTW & 10.45\% & \greybg{10.14\%} & 10.19\% & \greybg{9.69\%} & \greybg{9.29\%} & 9.30\% \\
    CVRPBLTW & 7.75\% & \greybg{7.70\%} & 7.47\% & \greybg{7.29\%} & 6.94\% & \greybg{6.92\%} \\
    OCVRPL & 3.44\% & \greybg{2.86\%} & 3.24\% & \greybg{2.69\%} & 2.31\% & \greybg{2.14\%} \\
    OCVRPBLTW & 10.50\% & \greybg{10.27\%} & 10.26\% & \greybg{9.80\%} & \greybg{9.22\%} & 9.35\% \\
    CVRPBTW & 7.41\% & \greybg{7.39\%} & 7.19\% & \greybg{6.92\%} & 6.74\% & \greybg{6.52\%} \\
    OCVRPBL & 7.34\% & \greybg{6.36\%} & 7.30\% & \greybg{6.26\%} & 5.41\% & \greybg{5.13\%} \\
    OCVRPLTW & 4.37\% & \greybg{3.22\%} & 3.97\% & \greybg{2.99\%} & 3.16\% & \greybg{2.50\%} \\
    \midrule
    \midrule
    Avg.Gap  & 4.97\% & \greybg{4.37\%} & 4.71\% & \greybg{4.16\%} & 3.93\% & \greybg{3.68\%} \\
    Best Solution & 0/16  & \greybg{16/16} & 0/16  & \greybg{16/16} & 2/16  & \greybg{14/16} \\
    \bottomrule[0.5mm]
    \end{tabular}%
  }
  \label{tab:multi_task_result_mtpomo_mvmoe_reld}%
\end{center}
\end{wraptable}%

We conduct experiments on larger-scale instances to validate CARM’s cross-scale scalability. As shown in Table \ref{tab:stl_large_scale}, the performance advantages of CARM in single-task learning become even more notable as the problem scale increases. This is particularly impressive with POMO~\citep{kwon2020pomo}, which achieves substantial improvements across all evaluated VRP variants. By improving constraint awareness capabilities, the CARM-enhanced solvers enable superior scalability without relying on large-scale training instances~\citep{zhou2024icam,zhou2023omni} or additional distance heuristics~\citep{gao2023elg,zhou2024icam,zhou2025L2R}. These findings provide valuable insights for the architectural design of neural routing solvers.

\subsection{Results with Multi-Task Learning}

\paragraph{Performance under Few-variant Training}

For multi-task learning, we first re-evaluate the three models discussed in Section \ref{sec:diff_embedding}. As shown in Table \ref{tab:multi_task_result_mtpomo_mvmoe_reld}, compared to the poor performance on unseen variants when solely applying FGE (as previously shown in Table \ref{tab:mtpomo_mvmoe_reld_OoD}), the three models equipped with the CARM-enhanced FGE achieve substantial improvements across both seen and unseen VRPs. Notably, MTPOMO~\citep{liu2024mtpomo} and MVMoE~\citep{zhou2024mvmoe} outperform the existing PRE-based version on all 16 variants. This demonstrates that CARM effectively mitigates the identical observation space issue induced by the FGE, enabling existing multi-task solvers to leverage the advantages of FGE and consistently achieve superior performance across diverse VRP variants.

\begin{table}[htbp]
  \centering
  \caption{Comparison of RF and CaDA with and without the CARM module across 24 VRPs (16 seen and 8 unseen variants). The last 8 variants with MB constraints are unseen, while the rest are seen.}
  \resizebox{\linewidth}{!}{
    \begin{tabular}{l|cccccccc}
    \toprule[0.5mm]
    Method & CVRP  & OCVRP & CVRPB & CVRPTW & CVRPL & OCVRPTW & OCVRPB & CVRPBL \\
    \midrule
    RF-PRE & \greybg{1.51\%} & 4.06\% & 3.95\% & 3.18\% & 1.83\% & 2.35\% & 4.21\% & 5.04\% \\
    RF-CARM & 1.53\% & \greybg{3.94\%} & \greybg{3.85\%} & \greybg{2.42\%} & \greybg{1.82\%} & \greybg{1.62\%} & \greybg{3.76\%} & \greybg{4.80\%} \\
    \midrule
    CaDA-PRE & 1.52\% & 4.00\% & 3.91\% & 3.15\% & 1.86\% & 2.35\% & 4.10\% & 5.00\% \\
    CaDA-CARM & \greybg{1.50\%} & \greybg{3.98\%} & \greybg{3.74\%} & \greybg{2.11\%} & \greybg{1.76\%} & \greybg{1.35\%} & \greybg{3.79\%} & \greybg{4.73\%} \\
    \midrule
    \midrule
          & CVRPLTW & OCVRPBTW & CVRPBLTW & OCVRPL & OCVRPBLTW & CVRPBTW & OCVRPBL & OCVRPLTW \\
    \midrule
    \midrule
    RF-PRE & 3.58\% & 2.04\% & 2.92\% & 4.05\% & 2.05\% & 2.62\% & 4.26\% & 2.35\% \\
    RF-CARM & \greybg{2.84\%} & \greybg{1.38\%} & \greybg{2.29\%} & \greybg{3.94\%} & \greybg{1.39\%} & \greybg{1.94\%} & \greybg{3.79\%} & \greybg{1.63\%} \\
    \midrule
    CaDA-PRE & 3.00\% & 1.44\% & 2.36\% & 4.05\% & 1.44\% & 2.00\% & 4.10\% & 1.75\% \\
    CaDA-CARM & \greybg{2.46\%} & \greybg{1.11\%} & \greybg{1.98\%} & \greybg{3.97\%} & \greybg{1.11\%} & \greybg{1.63\%} & \greybg{3.82\%} & \greybg{1.35\%} \\
    \midrule
    \midrule
          & CVRPMB & OCVRPMB & CVRPMBL & OCVRPMBL & CVRPMBTW & OCVRPMBTW & CVRPMBLTW & OCVRPMBLTW \\
    \midrule
    \midrule
    RF-PRE & 10.18\% & 18.90\% & 10.37\% & 18.87\% & 10.84\% & \greybg{8.73\%} & 10.81\% & \greybg{8.73\%} \\
    RF-CARM & \greybg{9.65\%} & \greybg{17.51\%} & \greybg{9.62\%} & \greybg{17.44\%} & \greybg{10.48\%} & 8.74\% & \greybg{10.65\%} & 8.84\% \\
    \midrule
    CaDA-PRE & 9.69\% & 17.38\% & \greybg{9.29\%} & 17.39\% & 11.53\% & 9.49\% & 11.45\% & 9.48\% \\
    CaDA-CARM & \greybg{9.34\%} & \greybg{16.66\%} & 9.40\% & \greybg{16.68\%} & \greybg{10.30\%} & \greybg{8.89\%} & \greybg{10.43\%} & \greybg{8.90\%} \\
    \bottomrule[0.5mm]
    \end{tabular}%
  }
  \label{tab:main_rf_cada_seen_unseen}%
\end{table}%

\paragraph{Performance under Many-variant Training}
To assess performance on a broader set of training problems, we evaluate RF~\citep{berto2024routefinder} and CaDA~\citep{li2025cada} trained on 16 variants, following their original settings. Table \ref{tab:main_rf_cada_seen_unseen} shows that both CARM-enhanced solvers achieve better overall performance on seen tasks than the original PRE, with CaDA~\citep{li2025cada} outperforming its original version across all seen variants. This advantage further extends to unseen 8 VRPs with MB constraints. These results confirm that the proposed CARM is an effective, model-agnostic module that improves neural solvers regardless of the number of training tasks. In Appendix \ref{app:all_results}, we present comprehensive results among PRE, FGE, and CARM across all five evaluated multi-task solvers.

\subsection{Ablation Study}

\begin{table}[htbp]
  \begin{minipage}[t]{0.47\linewidth}
    \centering
    \caption{Effects of FGE and CARM across 16 VRP variants.}
    \label{tab:ablation_mtpomo_summary}
    \resizebox{\linewidth}{!}{
      \begin{tabular}{cc|ccc|c}
      \toprule[0.5mm]
      State & \multirow{2}[2]{*}{CARM} & \multicolumn{3}{c|}{Avg.Gap} & Best \\
      Embedding &       & Seen (6) & Unseen (10) & All (16) & Solution \\
      \midrule
      PRE   & $\times$ & 2.86\% & 6.23\% & 4.97\% & 0/16 \\
      PRE   & $\checkmark$ & 2.46\% & 5.72\% & 4.50\% & 4/16 \\
      FGE   & $\times$ & 2.70\% & 6.46\% & 5.05\% & 0/16 \\
      FGE   & $\checkmark$ & \greybg{2.24\%} & \greybg{5.65\%} & \greybg{4.37\%} & \greybg{12/16} \\
      \bottomrule[0.5mm]
      \end{tabular}%
    }
  \end{minipage}
  \hfill 
  \begin{minipage}[t]{0.50\linewidth}
    \centering
    \caption{Ablation on parameter expansion techniques (CARM vs. standard linear projections).}
    \label{tab:summary_mtpomo_params}
    \resizebox{\linewidth}{!}{
      \begin{tabular}{l|cccc}
      \toprule[0.5mm]
      Strategy & Params. & $\Delta$ Params. & Avg.Gap & Best Sol. \\
      \midrule
      FGE-only   & 1.25M & 0      & 5.11\% & 0/16 \\
      FGE$+$PreLinear & 1.28M & 33.02K & 4.58\% & 1/16 \\
      FGE$+$PostLinear & 1.28M & 33.02K & 4.84\% & 0/16 \\
      FGE$+$CARM  & 1.27M & 17.79K & \greybg{4.37\%} & \greybg{15/16} \\
      \bottomrule[0.5mm]
      \end{tabular}%
    }
  \end{minipage}
\end{table}

To verify the contributions of FGE and CARM, we conduct an ablation study using MTPOMO~\citep{liu2024mtpomo} as a baseline. The results in Table \ref{tab:ablation_mtpomo_summary} reveal that applying FGE alone leads to superior results on seen problems but a clear performance degradation on unseen ones. Conversely, employing CARM alone effectively improves performance by improving the constraint awareness, even under the original PRE. Combining both components yields the best overall performance among all settings.

To further investigate the effectiveness of the CARM mechanism, we substitute our CARM module with two-layer linear projections ($d=128$) at two positions in MTPOMO~\citep{liu2024mtpomo}: one enhances $\mathbf{h}_{(C)}^{t}$ similar to ours (denoted as PreLinear), and the other enhances the attention output $\hat{\mathbf{h}}_{(C)}^{t}$ akin to ReLD~\citep{huang2025reld} (denoted as PostLinear). FGE and residual connection are retained in both variants. As detailed in Table \ref{tab:summary_mtpomo_params}, while both variants outperform the FGE-only strategy, FGE+CARM achieves the highest gains. This confirms that our CARM, not the increased parameter count, drives the performance improvements. Detailed results for all ablation studies are provided in Appendix \ref{app:ablation_studies}.

\section{Conclusion, Limitation, and Future Work}
\label{sec:conclusion}
In this work, we systematically revisit the existing state embedding generation mechanism in HELD-based neural routing solvers. We identify that the prevalent PRE restricts the observation space during attention computation, introducing a key bottleneck to achieving high-quality solutions. Through a detailed empirical analysis, we demonstrate the necessity of preserving a global observation space. Consequently, we propose a simple yet powerful CARM that enables the neural solver to fully leverage the global observation space and generate an efficient state embedding. Extensive results across two single-task and five multi-task solvers confirm that the CARM module consistently boosts baseline performance. Notably, solvers equipped with our CARM achieve substantial improvements in scaling to large-scale instances and in generalizing to unseen VRP variants.

\paragraph{Limitation and Future Work}
While our CARM consistently enhances the performance of various neural solvers, the current study focuses on model architectures. In the future, we aim to explore integrating constraint awareness into the training paradigm to further improve performance across diverse variants. A promising direction is to transition from the established practice of sampling $N$ trajectories with distinct starting nodes to a constraint-driven strategy for selecting starting nodes.

{\small

\bibliographystyle{unsrtnat} 

\bibliography{references}        
}
\clearpage

\appendix
\section{Related Work}
\label{app:related_work}
\subsection{HELD-based Single-Task Neural Routing Solvers}
NCO has emerged as the dominant paradigm for solving various VRP variants. As foundational work, the Attention Model (AM)~\citep{kool2019attention} adapted the Transformer architecture to canonical tasks such as TSP and CVRP, establishing the standard HELD framework. Building upon AM, POMO~\citep{kwon2020pomo} introduced a multi-start parallel decoding strategy that significantly improves solution quality, inspiring numerous subsequent architectural enhancements~\citep{xin2020stepwise,kim2022symnco,grinsztajn2023winner}.
Recent studies explore diverse strategies to scale these models, including local policy execution~\citep{gao2023elg}, cross-scale adaptive perception~\citep{zhou2024icam}, and search space reduction~\citep{li2021_L2D}. Alternatively, decomposition strategies integrate constructive models as sub-solvers to reduce the optimization complexity of the HELD framework~\citep{zheng2024udc,ye2023glop,hou2022tam}. Beyond scaling, the HELD architecture has been extensively adapted for complex, real-world constraints. Dedicated solvers explicitly address heterogeneous vehicle routing~\citep{li2022hcvrp,liu2025HCVRP}, the Pickup and Delivery TSP (PDTSP)~\citep{li2021pdtsp}, the CVRP with Backhauls (CVRPB)~\citep{wang2024cvrpb}, and min-max routing objectives~\citep{zheng2024dpn}. Similarly, specialized architectures such as MatNet~\citep{kwon2021matnet}, RRNCO~\citep{son2025rrnco}, and RADAR~\citep{yi2026radar} target asymmetric routing. Despite these advancements, existing HELD-based methods predominantly rely on PRE. This fundamentally restricts access to global contextual information, often yielding sub-optimal routing decisions.

\subsection{HELD-based Multi-Task Neural Routing Solvers}
To mitigate the computational overhead of training distinct models for each VRP variant, developing unified multi-task solvers has emerged as a prominent research direction. Initial approaches primarily focused on unified task formulation and efficient adaptation. For instance, MTPOMO~\citep{liu2024mtpomo} achieves zero-shot multi-task routing by formulating problem variants as combinations of base attributes, while RouteFinder~\citep{berto2024routefinder} utilizes Efficient Adapter Layers (EALs) to rapidly adapt pre-trained models via parameter-efficient fine-tuning. Subsequent studies have sought to enhance the capabilities of these models across various dimensions. To accommodate complex problem constraints, CaDA~\citep{li2025cada} incorporates constraint prompts and a dual-branch attention mechanism. Moreover, to address the inter-task interference that frequently plagues joint training, models like MVMoE~\citep{zhou2024mvmoe} and MoSES~\citep{pan2025moses} integrate Mixture-of-Experts (MoE) architectures, effectively decoupling parameters to mitigate task conflicts. Most recently, Scale-Net~\citep{liu2026scale} advanced this paradigm by extracting multi-scale structural representations to extend multi-task generalization to large-scale instances.
Despite these methodological strides, existing multi-task solvers remain fundamentally constrained by their reliance on PRE. Particularly in multi-task environments characterized by complex and intertwined constraints, the observation space induced by PRE is severely restricted. This restriction exacerbates the loss of global context during decision-making, ultimately limiting the models' routing capabilities.

\subsection{LEHD-based Neural Routing Solvers}
While HELD-based methods are highly efficient and widely successful, their reliance on static node embeddings fundamentally restricts their ability to capture dynamically evolving environment states during decoding~\citep{luo2023lehd}. In contrast, the LEHD architecture addresses this limitation by shifting the representational capacity to the decoder~\citep{luo2023lehd,drakulic2023bq}. However, this paradigm introduces significant trade-offs: the heavy decoder incurs substantial computational overhead and a large memory footprint~\citep{zheng2025mtl_kd, luo2024Boosting}. Furthermore, its reliance on fully supervised training demands massive optimal labels, rendering data acquisition prohibitively expensive. To alleviate the data acquisition and large-scale deployment challenges of the LEHD paradigm, recent research explores several enhanced strategies. For instance, SIL~\citep{luo2024Boosting} and SIGD~\citep{pirnay2024self} utilize self-generated, high-quality solutions for iterative training, bypassing the need for external labels. Additionally, MTL-KD~\citep{zheng2025mtl_kd} and TTPL~\citep{chen2025TTPL} leverage knowledge distillation and test-time alignment, respectively, to further reduce data dependency and enhance large-scale generalization. Notably, while the LEHD paradigm implicitly leverages FGE, it typically necessitates expensive supervised labels and expert knowledge. Within the more computationally efficient HELD framework, however, the integration of such a mechanism remains entirely unexplored.

\clearpage

\section{HELD-based Model Architecture}
\label{app:held_model_Architecture}
\subsection{Encoder}

The encoder $\bm{\theta}_{enc}$ consists of $L$ stacked attention layers. It initially projects the depot coordinates and node features into a unified d-dimensional space to obtain the initial embeddings $H^{(0)}$. Each encoder layer comprises two primary sub-layers: a Multi-Head Attention (MHA)\citep{vaswani2017attention} module and a Feed-Forward Network (FFN). These sub-layers are integrated using residual connections~\citep{he2016skip} and normalization~\citep{ioffe2015batchNorm,ulyanov2016instanceNorm}. For each layer $\ell$, let $H^{(\ell-1)} = \{\mathbf{h}_{i}^{(\ell-1)}\}_{i=0}^{n}$ be the input. The computation is defined as:
\begin{equation}
\widetilde{H}^{(\ell)} = \mathrm{Norm} \left( H^{(\ell-1)} + \mathrm{MHA}\left(H^{(\ell-1)}\right) \right),
\end{equation}
\begin{equation}
H^{(\ell)} = \mathrm{Norm} \left( \widetilde{H}^{(\ell)} + \mathrm{FFN}\left(\widetilde{H}^{(\ell)}\right) \right),
\end{equation}
where $\mathrm{FFN}(\cdot)$ denotes a position-wise network with ReLU activation. The generated $H^{(L)}$ remains unchanged during decoding.

\subsection{Decoder}
\label{append:model_decoder}
The decoder constructs a probability distribution over the unvisited nodes at each time step $t$ in an autoregressive manner. Prior to the sequential decoding phase, it pre-computes and caches the global key matrix $K$ and value matrix $V$ by projecting the final encoded node embeddings $H^{(L)}$ using learnable weight matrices $W_K$ and $W_V$:
\begin{equation}
K = H^{(L)}W_K, \quad V = H^{(L)}W_V.
\end{equation}
At each time step $t$, the decoder must be aware of both the current position and the remaining constraints. To achieve this, the embedding of the last visited node $\mathbf{h}_{\pi_{t-1}}^{(L)} \in \mathbb{R}^{d}$ is horizontally concatenated with the current dynamic constraint state $C_t$. This formulates a comprehensive context embedding $\mathbf{h}_{(C)}^{t} = [\mathbf{h}_{\pi_{t-1}}^{(L)}, C_t]$ (where $[\cdot,\cdot]$ denotes the horizontal concatenation operator), which serves as the query representation. Subsequently, the context embedding is mapped via a learnable matrix $W_Q$, and the updated state embedding $\hat{\mathbf{h}}_{(C)}^{t}$ is derived by applying a  MHA mechanism over the cached keys and values:
\begin{equation}
\hat{\mathbf{h}}_{(C)}^{t} = \mathrm{Attention}(W_Q\mathbf{h}_{(C)}^{t}, K, V,\mathcal{M}_t).
\end{equation}
Finally, to determine the next node, a compatibility score $\mathbf{u}^t=\{u^{t}_{i}\}_{i=0}^{n}$ is computed between the current state embedding and each candidate node's embedding. This score is clipped using a $\tanh$ function, scaled by a clipping hyperparameter $\xi$. Crucially, a masking function is applied by setting the score to $-\infty$ for already visited nodes $i \in \{\pi_{1:t-1}\}$, ensuring the generation of valid tours without duplicate visits. The scores are then passed through a Softmax function to obtain the final selection $p_{\bm{\theta}}(\pi_t=i \mid \pi_{1:t-1}, \mathcal{G}, C_{t})$ of selecting each node $i$.
\begin{equation}
u^{t}_{i} = \begin{cases} \xi \cdot \tanh\left(\frac{\hat{\mathbf{h}}_{(C)}^{t}(\mathbf{h}_{i}^{(L)})^\mathrm{T}}{\sqrt{d_k}}\right) & \text{if } i \not\in \{\pi_{1:t-1}\} \\ -\infty & \text{otherwise} \end{cases}.
\end{equation}
\begin{equation}
    p_{\bm{\theta}}(\pi_t=i \mid \pi_{1:t-1}, \mathcal{G}, C_{t}) = \mathrm{Softmax}(\mathbf{u}^t)
\end{equation}

\subsection{Training}

Within the framework of HELD-based neural solvers, the solution construction process can be formulated as a Markov Decision Process (MDP). The generated probability of a complete feasible solution $\bm{\pi}=(\pi_1, \pi_2,\pi_3,\dots, \pi_T)$ for instance $\mathcal{G}$ is calculated as: 
\begin{equation}
p_{\bm{\theta}}(\pi \mid \mathcal{G}) = \prod_{t=1}^{T} p_{\bm{\theta}}(\pi_t \mid \pi_{1:t-1}, \mathcal{G}, C_{t}) 
\end{equation}
where $T$ is the total number of construction steps, and $C_{t}$ indicates an optional constraint-relevant variables at step $t$ (e.g., remaining load for CVRP).

During training, the model parameters $\bm{\theta}$ are optimized to maximize the expected reward under the training data distribution via the REINFORCE algorithm~\citep{williams1992reinforce}. The gradient ascent with an approximation of the loss function can be written as:
\begin{equation}
    \nabla_\theta \mathcal{L} (\bm{\theta}) = \mathbb{E}_{p(\bm{\pi}| \mathcal{G},{\bm{\theta}})}[ 
    (f(\bm{\pi} | \mathcal{G})-b(\mathcal{G}))
    \nabla_{\bm{\theta}} \log p_{\bm{\theta}}\left(\bm{\pi} | \mathcal{G}\right) ],
    \label{eq:mean_loss}
\end{equation}
where the objective function $f(\bm{\pi} | \mathcal{G})$ represents the total reward (e.g., the negative value of tour length) of instance $\mathcal{G}$ given a specific solution $\bm{\pi}$, and $b(\mathcal{G})$ is the adopted baseline function. To obtain better performance, existing solvers generally leverage a shared baseline strategy by sampling $n$ parallel trajectories with distinct starting nodes~\citep{kwon2020pomo}:
\begin{equation}
    b(\mathcal{G}) = \frac{1}{n} \sum_{i=1}^n f(\bm{\pi}^{i} | \mathcal{G}).
\end{equation}

\clearpage
\section{Comparison of Gaussian Similarity}
\label{app:heatmaps}

\begin{figure}[htbp] 
    \centering

    \begin{subfigure}{0.245\textwidth}
        \centering
        \includegraphics[width=\textwidth]{figs/heat_map/cvrp_full_similarity.pdf}
        \caption{\centering CVRP under PRE}
        \label{app:subfig:heatmap_cvrp_full}
    \end{subfigure}\hfill
    \begin{subfigure}{0.245\textwidth}
        \centering
        \includegraphics[width=\textwidth]{figs/heat_map/cvrp_visited_similarity.pdf}
        \caption{\centering CVRP under FGE}
        \label{app:subfig:heatmap_cvrp_visited}
    \end{subfigure}\hfill
    \begin{subfigure}{0.245\textwidth}
        \centering
        \includegraphics[width=\textwidth]{figs/heat_map/cvrptw_full_similarity.pdf}
        \caption{\centering CVRPTW under PRE}
        \label{app:subfig:heatmap_cvrptw_full}
    \end{subfigure}\hfill
    \begin{subfigure}{0.245\textwidth}
        \centering
        \includegraphics[width=\textwidth]{figs/heat_map/cvrptw_visited_similarity.pdf}
        \caption{\centering CVRPTW under FGE}
        \label{app:subfig:heatmap_cvrptw_visited}
    \end{subfigure}

    \vspace{1em} %

    \begin{subfigure}{0.245\textwidth}
        \centering
        \includegraphics[width=\textwidth]{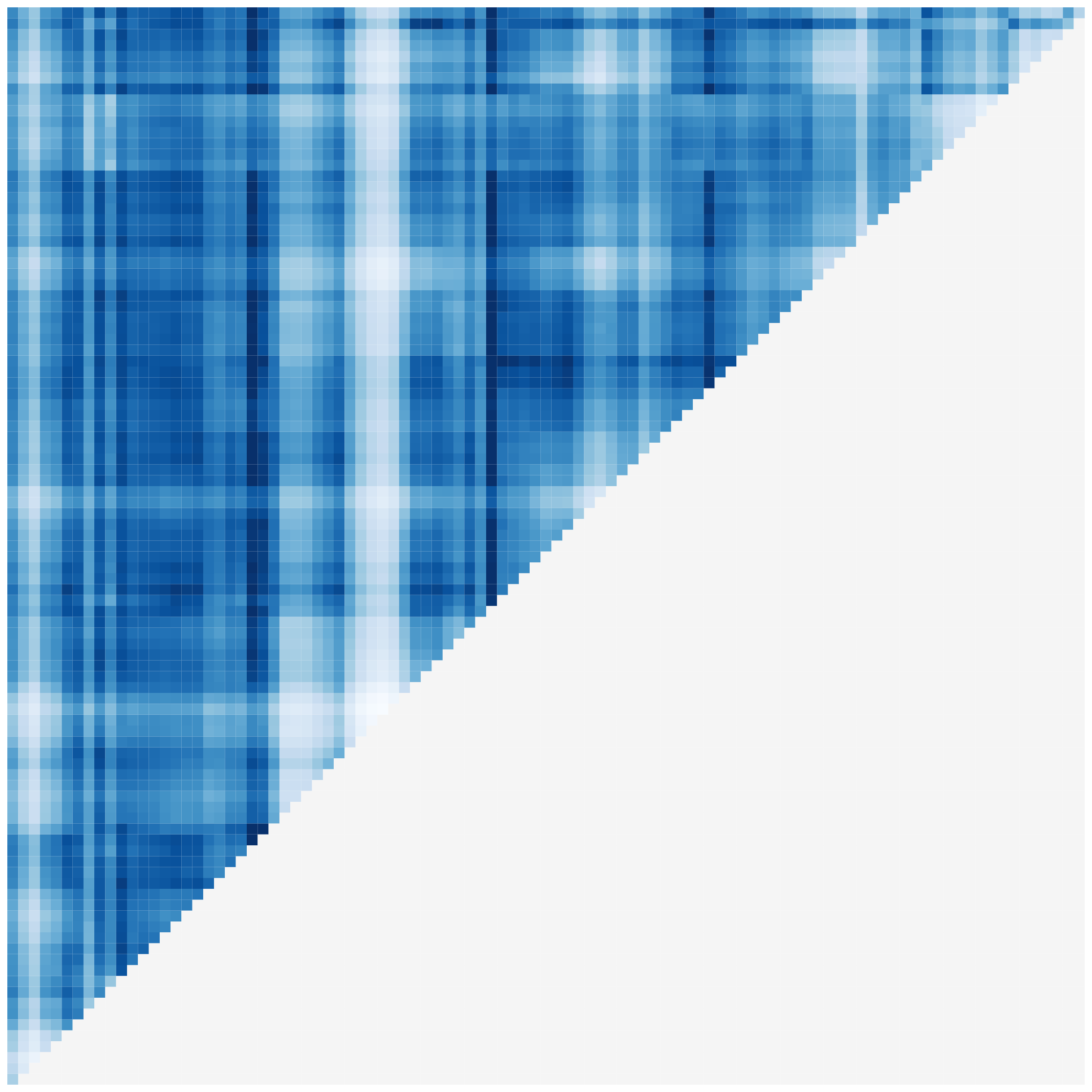}
        \caption{\centering CVRPB under PRE}
        \label{app:subfig:heatmap_cvrpb_full}
    \end{subfigure}\hfill
    \begin{subfigure}{0.245\textwidth}
        \centering
        \includegraphics[width=\textwidth]{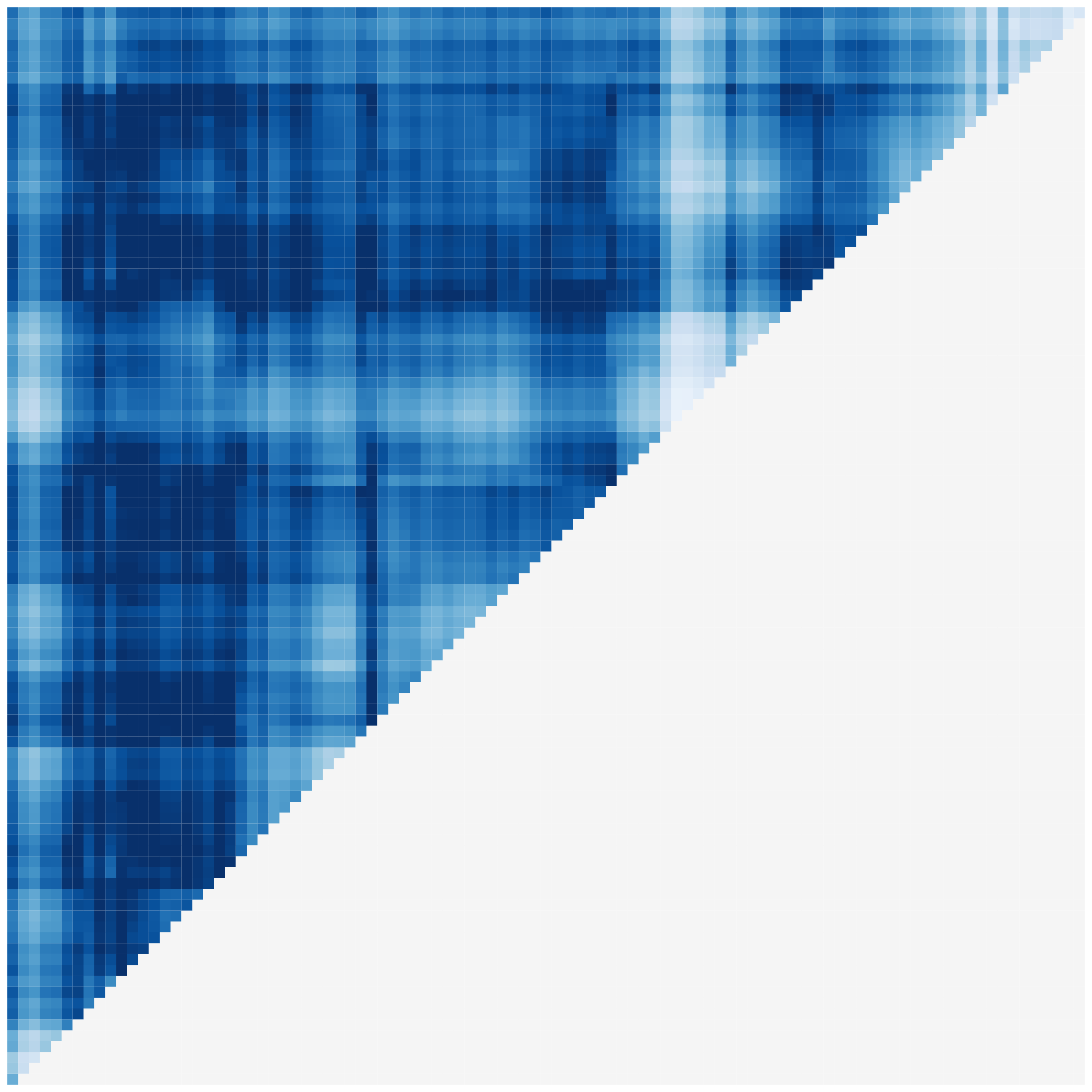}
        \caption{\centering CVRPB under FGE}
        \label{app:subfig:heatmap_cvrpb_visited}
    \end{subfigure}\hfill
    \begin{subfigure}{0.245\textwidth}
        \centering
        \includegraphics[width=\textwidth]{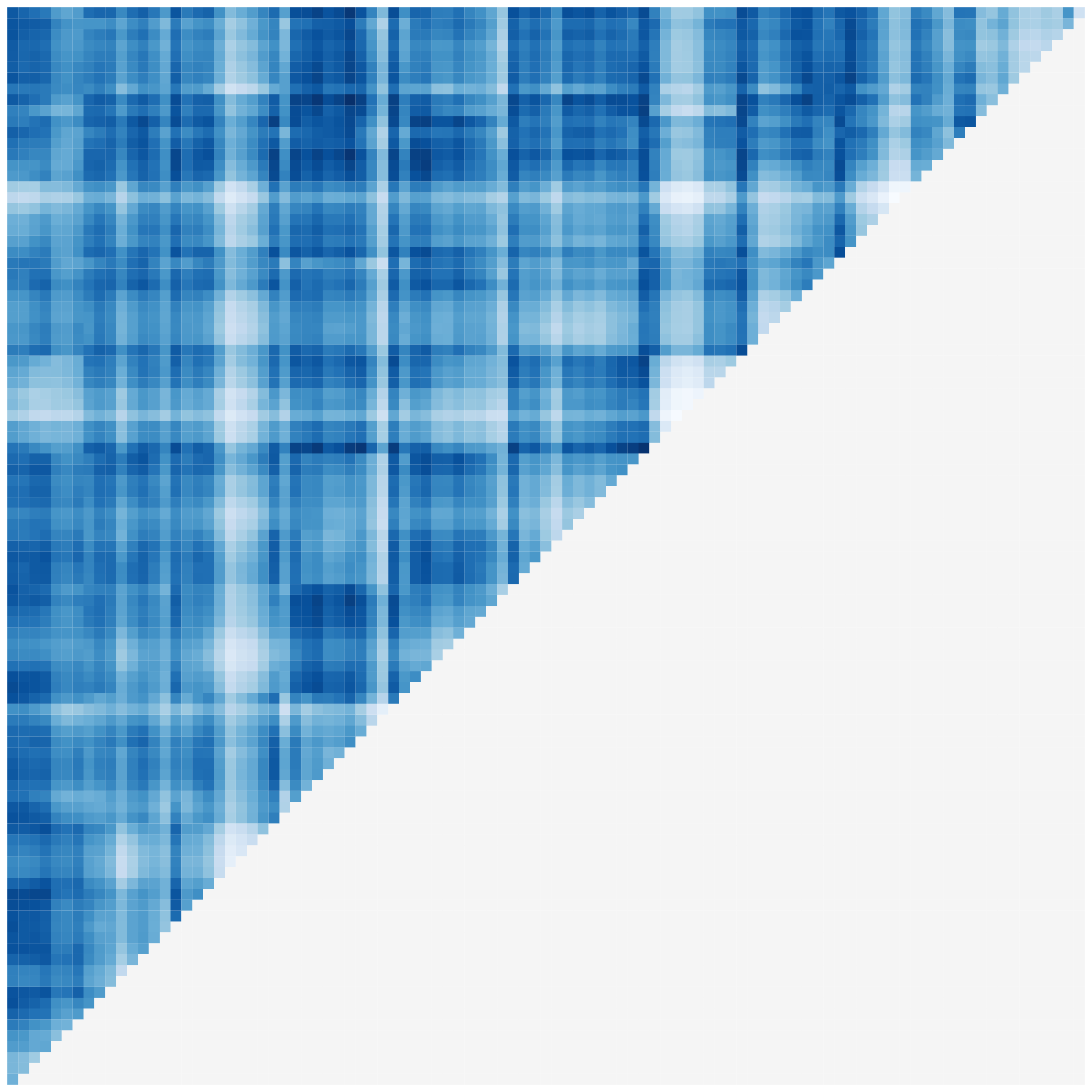}
        \caption{\centering CVRPL under PRE}
        \label{app:subfig:heatmap_cvrpl_full}
    \end{subfigure}\hfill
    \begin{subfigure}{0.245\textwidth}
        \centering
        \includegraphics[width=\textwidth]{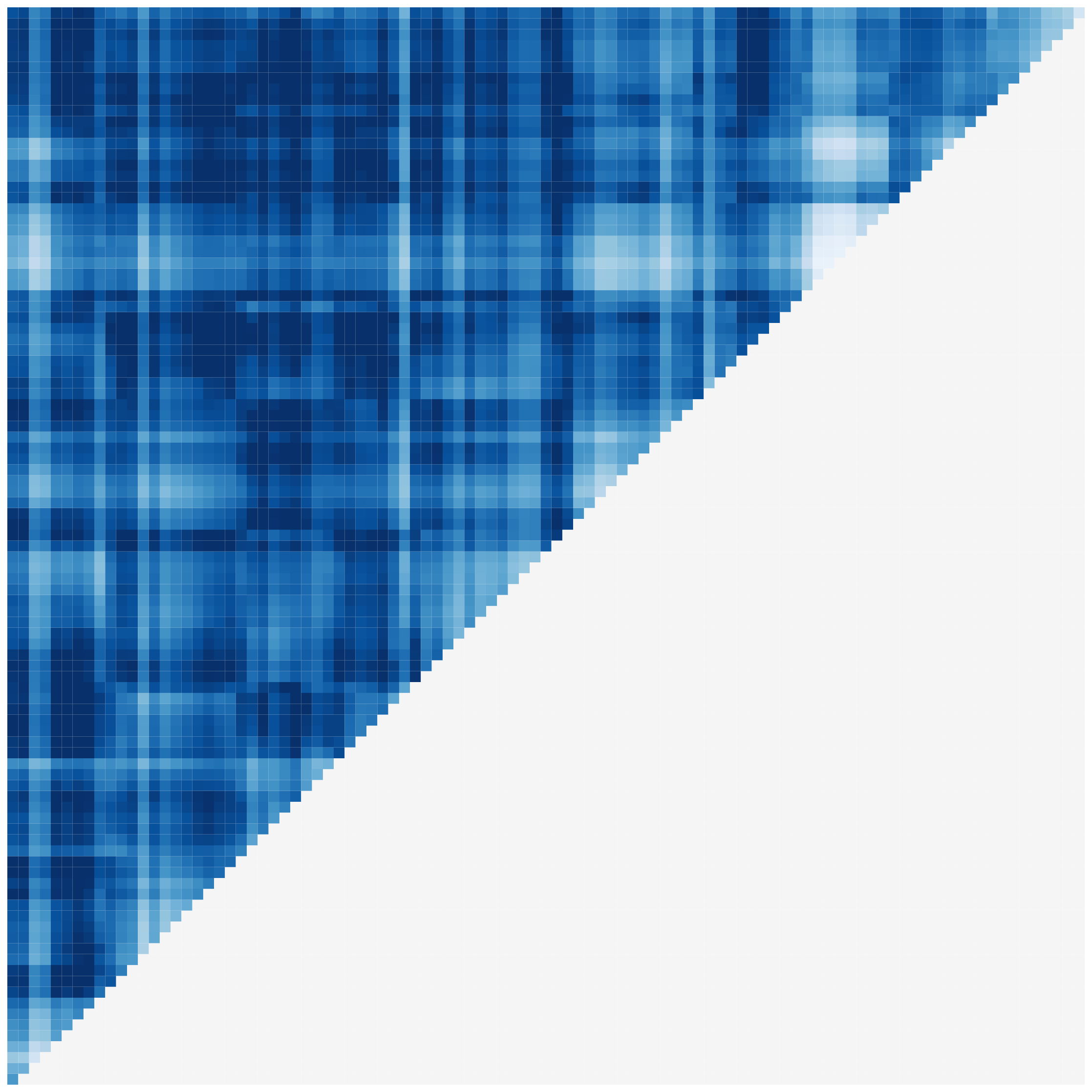}
        \caption{\centering CVRPL under FGE}
        \label{app:subfig:heatmap_cvrpl_visited}
    \end{subfigure}
    
    \vspace{1em} %

    \begin{subfigure}{0.245\textwidth}
        \centering
        \includegraphics[width=\textwidth]{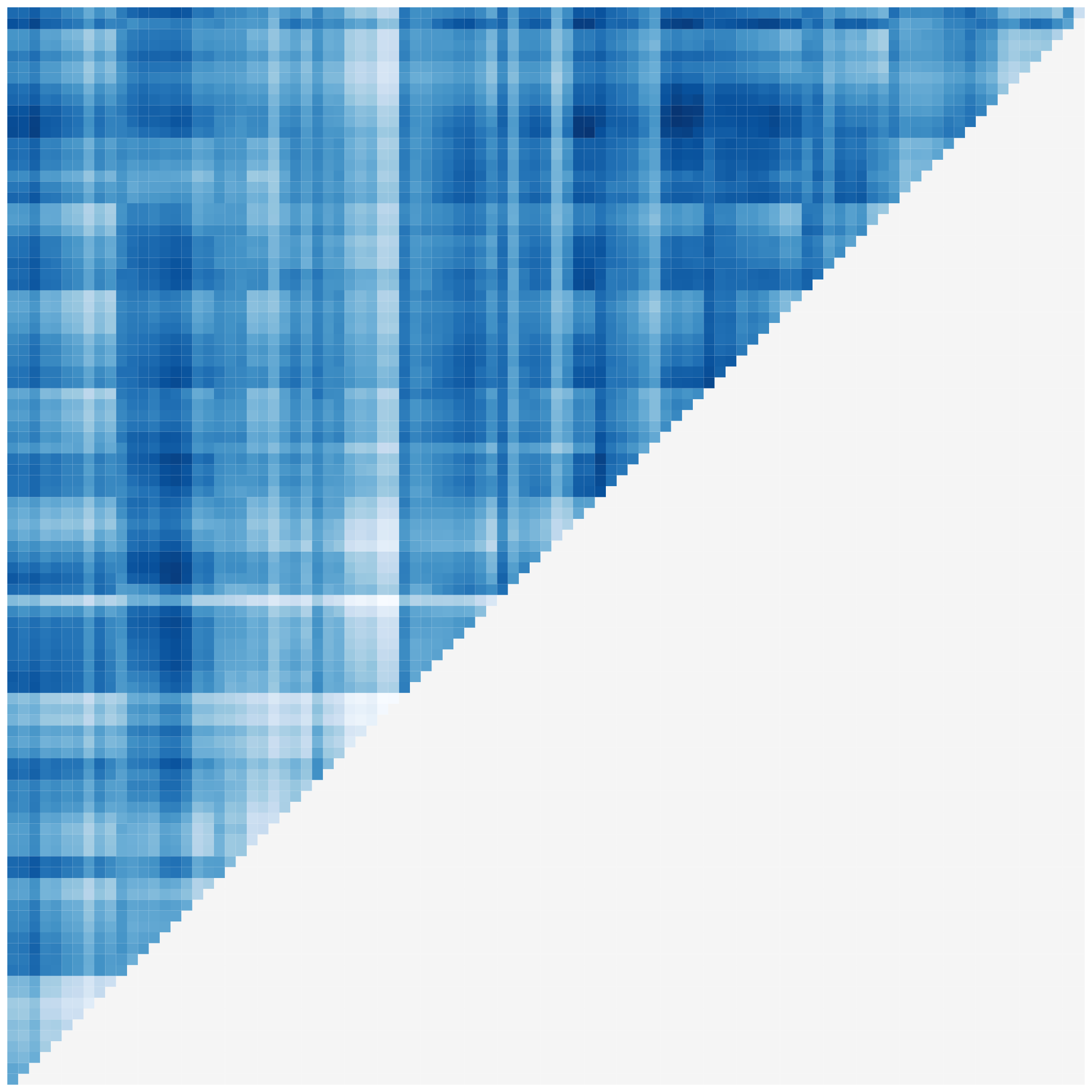}
        \caption{\centering OCVRP under PRE}
        \label{app:subfig:heatmap_ocvrp_full}
    \end{subfigure}\hfill
    \begin{subfigure}{0.245\textwidth}
        \centering
        \includegraphics[width=\textwidth]{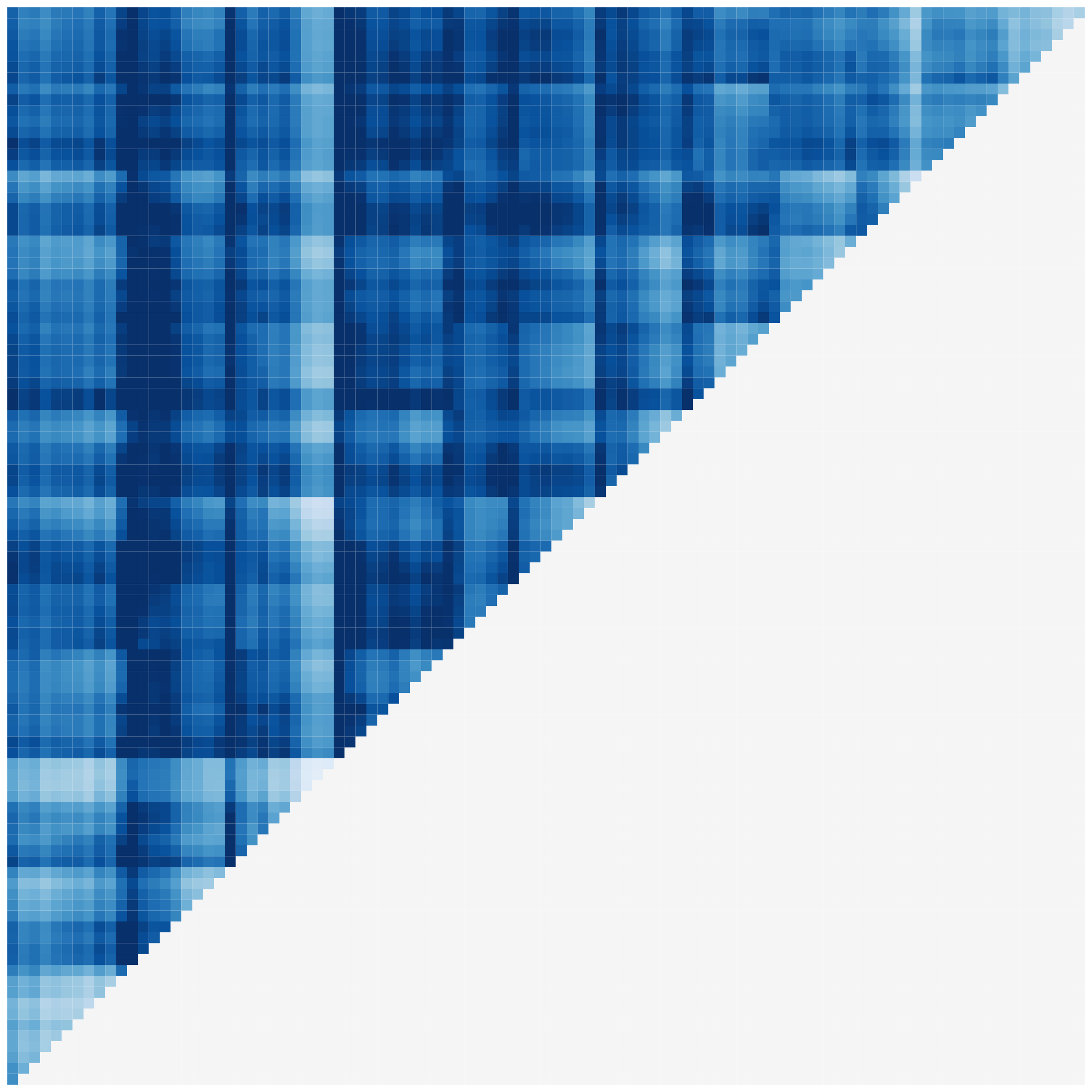}
        \caption{\centering OCVRP under FGE}
        \label{app:subfig:heatmap_ocvrp_visited}
    \end{subfigure}\hfill
    \begin{subfigure}{0.245\textwidth}
        \centering
        \includegraphics[width=\textwidth]{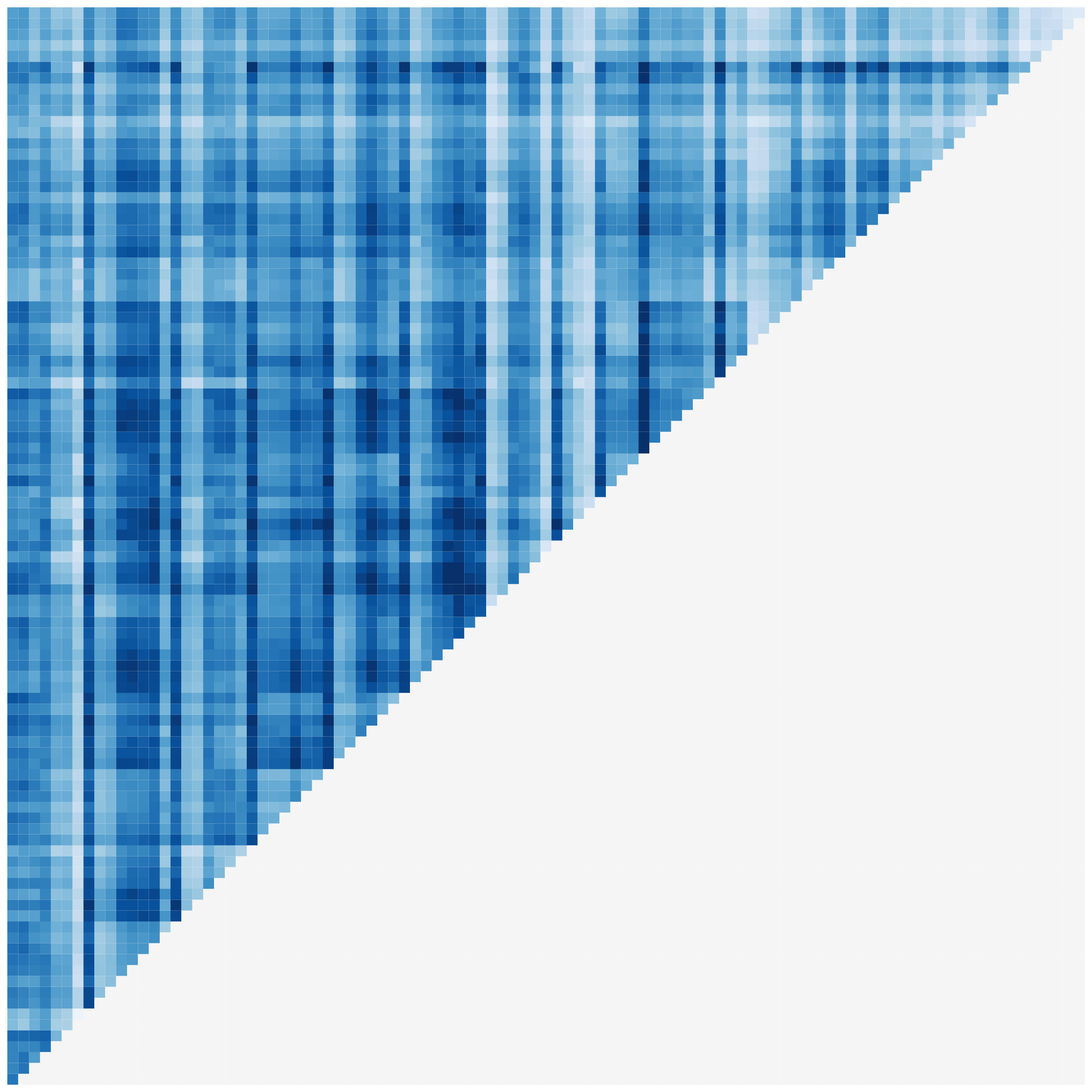}
        \caption{\centering OCVRPTW under PRE}
        \label{app:subfig:heatmap_ocvrptw_full}
    \end{subfigure}\hfill
    \begin{subfigure}{0.245\textwidth}
        \centering
        \includegraphics[width=\textwidth]{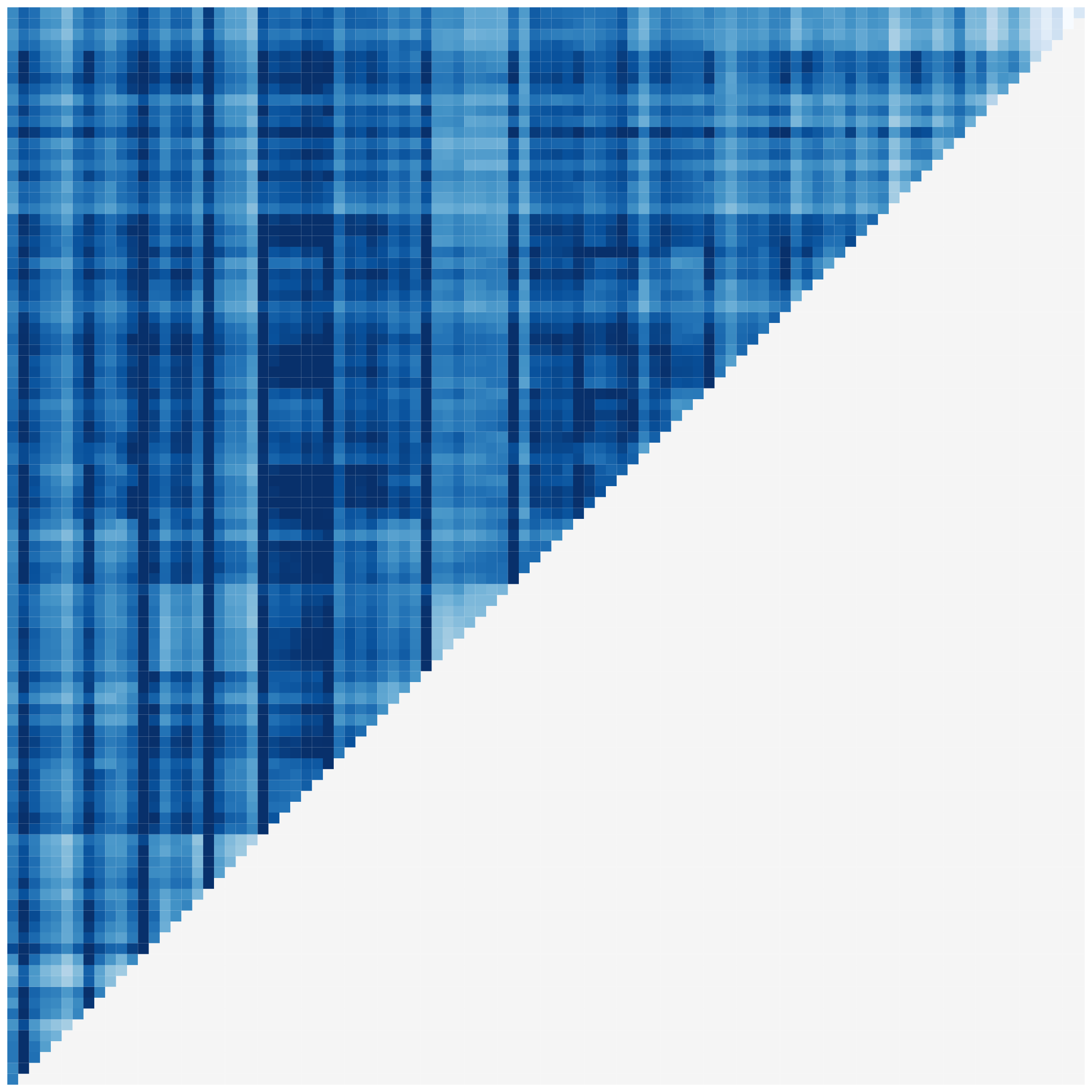}
        \caption{\centering OCVRPTW under FGE}
        \label{app:subfig:heatmap_ocvrptw_visited}
    \end{subfigure}

    \caption{Comparison of Gaussian similarity between the state embedding $\hat{\mathbf{h}}_{(C)}^t$ and node embeddings $H^{(L)}$ generated by PRE and FGE at each decoding step, evaluated using ReLD-MTL across various seen routing problems. It is noteworthy that darker shades indicate higher similarity.}
    \label{app:fig:heatmap_all}
\end{figure}

To investigate the impact of state embeddings on decision-making across different observation spaces, we take ReLD-MTL-PRE and ReLD-MTL-FGE as examples to compute the Gaussian similarity between the state embedding $\hat{\mathbf{h}}_{(C)}^t$ and the node embeddings $H^{(L)}=\{\mathbf{h}_i^{(L)}\}_{i=0}^{n}$ at each decoding step. The Gaussian similarity is defined as follows:
\begin{equation}
\begin{aligned}
    S(\hat{\mathbf{h}}_{(C)}^t, \mathbf{h}_i^{(L)}) &= \exp\left(-\gamma ||\hat{\mathbf{h}}_{(C)}^t - \mathbf{h}_i^{(L)}||^2\right) \\
    &= \exp\left(-\gamma \sum_{k=1}^{d} \left(\hat{\mathbf{h}}_{(C),k}^t - \mathbf{h}_{i,k}^{(L)}\right)^2\right)
\end{aligned}
\end{equation}
where $\mathbf{h}_i^{(L)}$ denotes the embedding of the $i$-th node, $d$ represents the embedding dimension, and the hyperparameter $\gamma$ is set to $2 \times 10^{-3}$. The value of $S(\hat{\mathbf{h}}_{(C)}^t, \mathbf{h}_i^{(L)})$ ranges from $[0, 1]$. A similarity of $1$ indicates that the two compared embeddings coincide exactly in the feature space, whereas a value approaching $0$ implies they are distant and lack relevance.

We visualize the Gaussian similarity using heatmaps, where darker colors indicate higher similarity values. We compare the two methods across six seen routing problems (i.e., CVRP, CVRPTW, CVRPB, CVRPL, OCVRP, and OCVRPTW). To align the heatmaps with the routing decisions, both axes are strictly ordered by the solution sequence generated by each method. For example, a route of $0 \rightarrow 5 \rightarrow 1 \rightarrow 3 \rightarrow 2 \rightarrow 4 \rightarrow 0$ (where $0$ is the depot) corresponds to an axis sequence of $5, 1, 3, 2, 4$. Because the number of depot returns varies among solutions, we ensure a fair comparison by plotting only the similarity matrix for the $n$ customer nodes. Furthermore, we mask out the similarities for already-visited nodes.

As illustrated in Figure \ref{app:fig:heatmap_all}, a comparative analysis of the similarity matrices reveals a consistent structural divergence between the two state embeddings. Across all six evaluated seen routing problems, the heatmaps corresponding to ReLD-MTL-FGE demonstrate a noticeably denser and darker color distribution compared to the baseline ReLD-MTL-PRE. Given that darker shades represent elevated Gaussian similarity values, this visual evidence indicates a stronger correlation between the current state embedding $\hat{\mathbf{h}}_{(C)}^t$ and the unvisited node embeddings $H^{(L)}$ within the FGE. From a mechanistic perspective, this pronounced similarity suggests that the state representations generated by FGE are more tightly aligned with the feature space of prospective nodes. Consequently, it can be inferred that FGE more effectively captures and integrates comprehensive global state information at each decoding step, thereby providing a richer and more accurate context to guide subsequent routing decisions.

\clearpage
\section{Setups of VRP Variants}
\label{app:setups_of_vrp_variants}
Current mainstream settings for VRP variants predominantly fall into two frameworks: one follows the experimental configuration of MVMoE\citep{zhou2024mvmoe}, while the other adopts the standards of RouteFinder\citep{berto2024routefinder}. Except for the settings of Capacity and Open Route, which remain consistent, the remaining constraints differ between the two frameworks. This section will elaborate on these constraint conditions in detail.

\paragraph{Capacity (C)}
Except for the depot, each customer node is associated with a specific demand. Before the completion of any single service route, the total demand loaded on a vehicle must not exceed its capacity. In both MVMoE and RouteFinder, following the configuration in AM \citep{kool2019attention}, the vehicle capacity is set to $\mathcal{C}=50$ for all VRP variants. Furthermore, the demand of each node is uniformly sampled from the discrete set $\{1, 2, \dots, 9\}$.

\paragraph{Open Route (O)}
Under the open route setting, vehicles need not return to the depot after service. Consequently, the final cost evaluation excludes the distance from the last visited customer back to the depot. 

\paragraph{Backhaul (B)}

The standard CVRP involves vehicles delivering goods from a central depot to customers with positive demands. The CVRP with Backhauls (CVRPB) extends this paradigm by requiring vehicles to not only service standard linehaul customers but also collect goods from backhaul customers for return to the depot. Customer demands are uniformly sampled from the discrete set $\{1, \dots, 9\}$. Under the MVMoE setting, 20\% of the nodes are randomly designated as backhaul customers with negative demands, while the remainder are linehaul customers; no precedence constraints are imposed between linehaul and backhaul visits. 
In the RouteFinder setting, each node is independently assigned a 20\% probability of being a backhaul customer. RouteFinder setting enforces a precedence constraint wherein all linehaul customers must be serviced prior to any backhaul customers within a single route, and demand values for both customer types remain positive.

\paragraph{Duration Limit (L)}
The Duration Limit constraint imposes an upper bound on the maximum length of any single vehicle route. Under the MVMoE setting, this limit is fixed at 3. Since all node coordinates are generated within a unit square $\mathcal{U}(0,1)$, the maximum round-trip distance to any single node is strictly bounded by $2\sqrt{2}$. Given that $2\sqrt{2} < 3$, this configuration inherently guarantees the existence of feasible solutions. In the RouteFinder setting, the duration limit is sampled from a uniform distribution $\mathcal{U}(2\max_i d_{0i}, l_{\max})$, where $d_{0i}$ denotes the distance from the depot to customer $i$, and $l_{\max} = 3$ serves as the predefined upper bound. For multi-depot variants, the term $\max_i d_{0i}$ is replaced by $\min_j \max_i d_{ij}$, where $i \in \{m, \dots, m + n\}$ represents the customer nodes, $j \in \{0, \dots, m\}$ indicates the depot nodes, $n$ specifies the problem size, and $m$ denotes the number of depots.

\paragraph{Time Windows (TW)}
The TW constraint requires vehicles to initiate service at node $v_i$ within a specific interval $[e_i, l_i]$ and incur a service time $s_i$. We employ two distinct generation strategies based on the experimental configuration. In the MVMoE setting, the depot's window and service time are fixed at $[0, 3]$ and $0$, respectively, while customer service times $s_i$ are set to 0.2. The window is defined as $[\max(e_0, \gamma_i - \delta_i), \min(l_0, \gamma_i + \delta_i)]$, where the center $\gamma_i \sim \mathcal{U}(e_0 + d_{0i}, l_0 - d_{i0} - s_i)$ ($d_{0i}$ is the depot-to-node distance), and the half-width $\delta_i \sim \mathcal{U}(0.1, 1.0)$. Under the RouteFinder setting, service times $s_i$ and window lengths $t_i$ are sampled from $[0.15, 0.18]$ and $[0.18, 0.2]$, respectively. The time window is formulated as $[e_i, e_i + t_i]$, with the start time $e_i = (1 + (\beta_i - 1) \cdot u_i) \cdot d_{max}$, where $u_i \sim \mathcal{U}(0, 1)$, the upper bound factor $\beta_i = \frac{t_{max} - s_i - t_i}{d_{max}} - 1$, and $d_{max}$ denotes the maximum distance from any depot to $v_i$.

\paragraph{Mixed Backhaul(MB)}
Introduced within the RouteFinder setting, the MB constraint relaxes the strict precedence rules of traditional backhaul routing. Specifically, while the standard Backhaul constraint dictates that all linehaul nodes must be visited strictly prior to any backhaul nodes, the MB formulation permits the interleaving of linehaul and backhaul nodes within a single vehicle route.

\clearpage
\section{Feature-wise Linear Modulation}
\label{app:film}
Feature-wise Linear Modulation (FiLM)\citep{perez2018film} serves as a versatile conditioning mechanism designed to modulate the intermediate feature maps of a neural network. The fundamental principle involves applying an affine transformation to feature activations, conditioned on external signals such as textual descriptions, metadata, or cross-modal features. Due to its high parameter efficiency and minimal computational overhead, FiLM facilitates a highly flexible and effective integration of multimodal information.

Formally, let $x \in \mathbb{R}^{C \times H \times W}$ denote the feature activations of a specific layer, where $C$, $H$, and $W$ represent the number of channels, height, and width, respectively. For the $c$-th channel $x_c$, the FiLM operation applies a linear modulation defined as 
\begin{equation}
    \mathrm{FiLM}(x_c; \gamma_c, \beta_c) = \gamma_c \cdot x_c + \beta_c 
\end{equation}

In this formulation, the scaling factor $\gamma_c$ regulates the magnitude or relative importance of the feature channel, while the shifting factor $\beta_c$ adjusts the activation threshold by applying a bias to the feature values.

Rather than being learned as static weights, these modulation parameters are dynamically produced by a conditioning network $g(\cdot)$ based on an external input $z$, such that $\gamma, \beta = g(z)$. In practical implementations, $g(\cdot)$ is typically instantiated as a Multi-Layer Perceptron (MLP) with an output dimension of $2C$, which predicts a unique pair of $(\gamma_c, \beta_c)$ for each channel corresponding to the specific context provided by $z$.

Beyond its mathematical simplicity, FiLM enables sophisticated conditional logic within the latent space. By manipulating $\gamma_c$, the model performs implicit feature selection, effectively suppressing irrelevant channels ($\gamma_c \to 0$) or amplifying task-critical features. Concurrently, the bias term $\beta_c$ facilitates information injection by shifting the activation baseline independently of the primary input, thereby allowing external prior knowledge to directly influence the feature representations. This dynamic adaptability is particularly potent in tasks requiring fine-grained reasoning and contextual awareness.

\section{Oracle Solvers Settings}
\label{app:datasets}
To ensure fair comparisons, the configurations of the classical heuristic solvers are strictly aligned with the evaluation protocols of their respective neural baselines. Specifically, LKH3\citep{LKH3} executes 10,000 trials in a single run, and the search time limit for OR-Tools\citep{ortools} is restricted to 400 seconds. For HGS \citep{HGS}, the termination criterion depends on the comparison target: it is set to a maximum of 20,000 iterations without improvement when compared against MTPOMO \citep{liu2024mtpomo}, MVMoE \citep{zhou2024mvmoe}, and ReLD \citep{huang2025reld}, whereas a 20-second time limit is imposed when evaluated alongside RouteFinder \citep{berto2024routefinder} and CaDA \citep{li2025cada}. Furthermore, for large-scale instances, we utilize the dataset introduced by MTL-KD \citep{zheng2025mtl_kd} and employ PyVRP\citep{wouda2024pyvrp} as the classical baseline. To account for the increased complexity, the runtime limits for PyVRP are dynamically scaled to 300, 600, and 1200 seconds for problem sizes of $N=200$, $500$, and $1000$, respectively.

\clearpage
\section{Training Settings}
\label{app:training_settings}

All retrained baselines in this study strictly follow the hyperparameter configurations from their respective original publications\footnote{For MTPOMO~\citep{liu2024mtpomo}, we use the implementation provided by MVMoE~\citep{zhou2024mvmoe}, ensuring consistency across all three models.}. The shared hyperparameter settings are detailed in Table~\ref{tab:hyperparams_mtpomo_mvmoe_reld} and Table~\ref{tab:hyperparams_rf_cada}. For further details regarding model-specific implementations, readers are referred to the original papers.

\begin{table}[htbp]
  \centering
  \caption{Relevant hyperparameter configurations for MTPOMO\citep{liu2024mtpomo}, MVMoE\citep{zhou2024mvmoe}, and ReLD\citep{huang2025reld}.}
    \begin{tabular}{ll}
    \toprule
    Hyperparameter & Value \\
    \midrule
    \textbf{Model} &  \\
    Embedding dimension & 128 \\
    Number of attention heads & 8 \\
    Number of encoder layers & 6 \\
    Feedforward hidden dimension & 512 \\
    Tanh clipping ξ & 10.0  \\
    \midrule
    \textbf{Training} &  \\
    Batch size & 128 \\
    Train data per epoch & 20000 \\
    Optimizer &  Adam \\
    Learning rate (LR) & 1e-4 \\
    Weight decay & 1e-6 \\
    LR scheduler & MultiStepLR \\
    LR milestones & [4501] \\
    LR gamma & 0.1 \\
    Training epochs & 5000 \\
    Number of tasks used for training & 6 \\
    \bottomrule
    \end{tabular}%
  \label{tab:hyperparams_mtpomo_mvmoe_reld}%
\end{table}%

\begin{table}[htbp]
  \centering
  \caption{Relevant hyperparameter configurations for RouteFinder\citep{berto2024routefinder},CaDA\citep{li2025cada}}
    \begin{tabular}{ll}
    \toprule
    Hyperparameter & Value \\
    \midrule
    \textbf{Model} &  \\
    Embedding dimension & 128 \\
    Number of attention heads & 8 \\
    Number of encoder layers & 6 \\
    Feedforward hidden dimension & 512 \\
    Tanh clipping ξ & 10.0  \\
    \midrule
    \textbf{Training} &  \\
    Batch size & 256 \\
    Train data per epoch & 100000 \\
    Optimizer &  AdamW \\
    Learning rate (LR) & 3e-4 \\
    Weight decay & 1e-6 \\
    LR scheduler & MultiStepLR \\
    LR milestones & [270,295] \\
    LR gamma & 0.1 \\
    Gradient clip value & 1 \\
    Training epochs & 300 \\
    Number of tasks used for training & 16 \\
    \bottomrule
    \end{tabular}%
  \label{tab:hyperparams_rf_cada}%
\end{table}%

\clearpage
\section{Implementation Details of CARM Across Various Solvers}
\label{app:carm}
\subsection{ReLD-CARM}
In the decoder of ReLD \citep{huang2025reld}, standard attention computations are first performed:
\begin{equation}
    Q=\mathbf{h}_{(C)}^tW_Q, \quad K=H^{(L)}W_K,\quad V=H^{(L)}W_V
\end{equation}
\begin{equation}
    \mathbf{h}_{(A)}^{t} = \mathrm{Attention}(Q,K,V,\mathcal{M}_t)
\end{equation}

where $\mathbf{h}_{(C)}^{t}=[\mathbf{h}_{\pi_{t-1}}^{(L)}, C_t]$, with $\mathbf{h}_{\pi_{t-1}}^{(L)}$ denoting the embedding of the last visited node and $C_t$ representing the current constraint state. To enhance constraint awareness, ReLD improves upon the attention output $\mathbf{h}^t_{(A)}$ by incorporating an identity mapping function $\mathrm{IDT}(\cdot)$ and a Feed-Forward Network (FFN), yielding the PRE-based state embedding $\hat{\mathbf{h}}_{(C_p)}^t$. The specific formulations are as follows:
\begin{equation}
    \bar{\mathbf{h}}_{(A)}^{t} = \mathbf{h}_{(A)}^{t} + \mathrm{IDT}(\mathbf{h}_{\pi_{t-1}}^{(L)})
\end{equation}
\begin{equation}
    \mathrm{IDT}(\mathbf{h}_{\pi_{t-1}}^{(L)}) = \mathbf{h}_{\pi_{t-1}}^{(L)} + W_{idt}C_t
\end{equation}
\begin{equation}
    \hat{\mathbf{h}}_{(C_p)}^t = \mathrm{FF}(\bar{\mathbf{h}}_{(A)}^{t})+\bar{\mathbf{h}}_{(A)}^{t}
\end{equation}

Building upon this framework, we introduce the CARM module. Adhering to the principle of minimal architectural modification, our adaptations are lightweight and seamlessly integrated. Specifically, rather than redesigning the decoder, we merely replace $\mathbf{h}_{(C)}$ with the Constraint-Conditioned Modulation output $\mathbf{h}^t_{(M)}$ (detailed in Section \ref{sec:carm}). Subsequently, during the attention computation, we apply masking exclusively to the visited nodes. Then, prior to the FFN computation, we incorporate an additional residual connection for $\mathbf{h}^t_{(M)}$ while keeping the original identity mapping structure intact. The final FGE-based state embedding is formulated as:
\begin{equation}
    \bar{\mathbf{h}}_{(A)}^{'t} = \mathbf{h}_{(A)}^{t} + \mathbf{h}^t_{(M)} +\mathrm{IDT}(\mathbf{h}_{\pi_{t-1}}^{(L)})
\end{equation}
\begin{equation}
   \hat{\mathbf{h}}_{(C_f)}^t = \mathrm{FF}(\bar{\mathbf{h}}_{(A)}^{'t})+\bar{\mathbf{h}}_{(A)}^{'t} 
\end{equation}

We present an ablation study in Table \ref{tab:ablation_reld} to demonstrate that the performance boost in ReLD-CARM is largely attributed to the CARM module instead of the IDT. Notably, the variant without IDT maintains an average performance on par with the ReLD-CARM, underscoring the superiority of our CARM module over ReLD's original constraint mapping mechanism.

\subsection{Other Solvers with CARM}
Except for ReLD, the other solvers integrate the CARM module following the procedure detailed in Section \ref{sec:methodology}. Specifically, when computing $Q$, we replace $\mathbf{h}_{(C)}^t$ with the Constraint-Conditioned Modulation output $\mathbf{h}^t_{(M)}$. Subsequently, we apply masking exclusively to the visited nodes during the attention computation, and incorporate a residual connection between $\mathbf{h}^t_{(M)}$ and the attention output.

\clearpage
\section{Analysis of the Computational Cost of the CARM Module}
\label{app:inference_time_analysis}
We evaluate the inference time and optimality gap of POMO\citep{kwon2020pomo} and ReLD\citep{huang2025reld} in single-task large-scale experiments on CVRPTW and CVRPB, as summarized in Table \ref{tab:single_task_time_test}. For problem scales of $N=200$ and $500$, incorporating the CARM module yields steady improvements in the optimality gap with only marginal variations in inference time (with time overheads not exceeding 1.5 seconds). Notably, at the larger scale of $N=1000$, the integration of CARM consistently decreases the total inference time across both baselines. This acceleration occurs because the enhanced decision quality reduces the total number of decoding steps required to construct the routes.

\begin{table}[htbp]
  \centering
  \caption{Scalability and inference time evaluation on larger-scale instances of CVRPTW and CVRPB. All models are trained with single-task learning, and the training scale is fixed to $100$.}
 
    \begin{tabular}{l|cccccc}
    \toprule[0.5mm]
    \multicolumn{1}{c|}{\multirow{2}[2]{*}{Method}} & \multicolumn{2}{c}{CVRPTW200} & \multicolumn{2}{c}{CVRPTW500} & \multicolumn{2}{c}{CVRPTW1000} \\
          & Gap   & Time  & Gap   & Time  & Gap   & Time \\
    \midrule
    POMO  & 9.75\% & 6.92s & 29.74\% & 92.92s & 69.64\% & 771.98s \\
    POMO-CARM & \greybg{7.41\%} & 7.53s & \greybg{16.13\%} & 90.29s & \greybg{22.07\%} & 674.35s \\
    \midrule
    \midrule
    ReLD  & 7.29\% & 7.70s & 14.29\% & 94.40s & 19.00\% & 695.90s \\
    ReLD-CARM & \greybg{6.42\%} & 8.30s & \greybg{12.93\%} & 95.74s & \greybg{16.40\%} & 687.96s \\
    \midrule
          & \multicolumn{2}{c}{CVRPB200} & \multicolumn{2}{c}{CVRPB500} & \multicolumn{2}{c}{CVRPB1000} \\
          & Gap   & Time  & Gap   & Time  & Gap   & Time \\
    \midrule
    POMO  & 0.39\% & 4.38s & 4.73\% & 54.10s & 11.38\% & 429.80s \\
    POMO-CARM & \greybg{-0.99\%} & 5.12s & \greybg{0.07\%} & 54.39s & \greybg{2.69\%} & 402.16s \\
    \midrule
    \midrule
    ReLD  & -1.29\% & 5.01s & 0.74\% & 57.75s & 6.62\% & 424.92s \\
    ReLD-CARM & \greybg{-1.90\%} & 5.48s & \greybg{-1.61\%} & 58.51s & \greybg{1.17\%} & 415.07s \\
    \bottomrule[0.5mm]
    \end{tabular}%

  \label{tab:single_task_time_test}%
\end{table}%

Furthermore, Table \ref{tab:multi_task_time_test} reports the optimality gaps and inference times for MTPOMO, MVMoE, and ReLD-MTL across 16 VRP variants, revealing a favorable trade-off between optimality and efficiency. Across the evaluated tasks, the CARM module narrows the average optimality gap for all baselines and yields the best solutions in most test cases. Crucially, this consistent enhancement in performance incurs a minimal computational overhead. Specifically, CARM introduces an average latency penalty of only 0.24s for MTPOMO, 1.09s for MVMoE, and 0.83s for ReLD-MTL. Such a tightly bound time variation confirms that the improved decision quality does not lead to a significant degradation in inference speed.

\begin{table}[htbp]
  \centering
  \caption{Comparison of optimality gaps and inference times for solvers with and without the CARM module across 16 VRPs (6 seen and 10 unseen variants). Variants marked with an asterisk ($^*$) indicate seen tasks.}
  \resizebox{\linewidth}{!}{
    \begin{tabular}{l|cccc|cccc|cccc}
    \toprule[0.5mm]
    \multicolumn{1}{c|}{\multirow{3}[2]{*}{Problem}} & \multicolumn{4}{c|}{MTPOMO}   & \multicolumn{4}{c|}{MVMoE}    & \multicolumn{4}{c}{ReLD-MTL} \\
          & \multicolumn{2}{c}{PRE} & \multicolumn{2}{c|}{CARM} & \multicolumn{2}{c}{PRE} & \multicolumn{2}{c|}{CARM} & \multicolumn{2}{c}{PRE} & \multicolumn{2}{c}{CARM} \\
          & Gap   & Time  & Gap   & Time  & Gap   & Time  & Gap   & Time  & Gap   & Time  & Gap   & Time \\
    \midrule
    CVRP* & 1.85\% & 7.06s & \greybg{1.59\%} & 7.99s & 1.73\% & 12.83s & \greybg{1.54\%} & 14.30s & 1.42\% & 8.11s & \greybg{1.34\%} & 9.15s \\
    OCVRP* & 3.46\% & 6.72s & \greybg{2.87\%} & 7.06s & 3.21\% & 12.46s & \greybg{2.77\%} & 13.45s & 2.32\% & 7.80s & \greybg{2.11\%} & 8.63s \\
    CVRPB* & 1.67\% & 5.89s & \greybg{1.11\%} & 5.91s & 1.37\% & 11.90s & \greybg{1.00\%} & 12.59s & 0.90\% & 6.46s & \greybg{0.67\%} & 7.03s \\
    CVRPTW* & 5.31\% & 7.34s & \greybg{4.54\%} & 7.59s & 4.93\% & 12.81s & \greybg{4.39\%} & 13.77s & 4.56\% & 8.32s & \greybg{4.09\%} & 9.15s \\
    CVRPL* & 0.48\% & 8.59s & \greybg{0.20\%} & 8.86s & 0.32\% & 15.27s & \greybg{0.16\%} & 16.30s & 0.02\% & 9.63s & \greybg{-0.04\%} & 10.54s \\
    OCVRPTW* & 4.41\% & 8.16s & \greybg{3.14\%} & 8.31s & 3.94\% & 15.02s & \greybg{2.88\%} & 16.09s & 3.10\% & 9.15s & \greybg{2.49\%} & 10.03s \\
    \midrule
    OCVRPB & 7.34\% & 6.09s & \greybg{6.25\%} & 6.20s & 7.24\% & 12.33s & \greybg{6.19\%} & 13.36s & 5.36\% & 6.77s & \greybg{5.09\%} & 7.34s \\
    CVRPBL & 1.79\% & 7.14s & \greybg{1.11\%} & 7.46s & 1.47\% & 12.97s & \greybg{0.98\%} & 13.95s & 1.01\% & 8.19s & \greybg{0.65\%} & 9.02s \\
    CVRPLTW & 1.92\% & 6.31s & \greybg{1.20\%} & 6.37s & 1.55\% & 12.45s & \greybg{1.00\%} & 13.20s & 1.17\% & 6.93s & \greybg{0.71\%} & 7.48s \\
    OCVRPBTW & 10.45\% & 7.78s & \greybg{10.14\%} & 8.02s & 10.19\% & 15.40s & \greybg{9.69\%} & 16.56s & \greybg{9.29\%} & 8.39s & 9.30\% & 9.38s \\
    CVRPBLTW & 7.75\% & 9.24s & \greybg{7.70\%} & 9.48s & 7.47\% & 15.92s & \greybg{7.29\%} & 17.12s & 6.94\% & 10.33s & \greybg{6.92\%} & 11.14s \\
    OCVRPL & 3.44\% & 6.41s & \greybg{2.86\%} & 6.49s & 3.24\% & 12.89s & \greybg{2.69\%} & 13.87s & 2.31\% & 6.99s & \greybg{2.14\%} & 7.60s \\
    OCVRPBLTW & 10.50\% & 7.46s & \greybg{10.27\%} & 7.77s & 10.26\% & 15.22s & \greybg{9.80\%} & 16.53s & \greybg{9.22\%} & 8.14s & 9.35\% & 9.14s \\
    CVRPBTW & 7.41\% & 8.36s & \greybg{7.39\%} & 8.60s & 7.19\% & 15.44s & \greybg{6.92\%} & 16.57s & 6.74\% & 9.45s & \greybg{6.52\%} & 10.29s \\
    OCVRPBL & 7.34\% & 8.08s & \greybg{6.36\%} & 8.40s & 7.30\% & 16.02s & \greybg{6.26\%} & 17.37s & 5.41\% & 8.86s & \greybg{5.13\%} & 9.80s \\
    OCVRPLTW & 4.37\% & 7.86s & \greybg{3.22\%} & 7.94s & 3.97\% & 15.94s & \greybg{2.99\%} & 17.17s & 3.16\% & 8.35s & \greybg{2.50\%} & 9.33s \\
    \midrule
    \midrule
    Avg.  & 4.97\% & 7.41s & \greybg{4.37\%} & 7.65s & 4.71\% & 14.05s & \greybg{4.16\%} & 15.14s & 3.93\% & 8.24s & \greybg{3.68\%} & 9.07s \\
    Best Solution & \multicolumn{2}{c}{0/16} & \multicolumn{2}{c|}{\greybg{16/16}} & \multicolumn{2}{c}{0/16} & \multicolumn{2}{c|}{\greybg{16/16}} & \multicolumn{2}{c}{2/16} & \multicolumn{2}{c}{\greybg{14/16}} \\
    \bottomrule[0.5mm]
    \end{tabular}%
  }
  \label{tab:multi_task_time_test}%
\end{table}%

\clearpage
\section{Ablation Studies}
\label{app:ablation_studies}

\subsection{Ablation Study on FGE and CARM}
To verify the individual contributions of FGE and CARM, we conduct a detailed ablation study based on MTPOMO~\citep{liu2024mtpomo}. The results in Table \ref{tab:ablation_mtpomo} reveal distinct effects for the two components. Applying FGE alone improves route quality on seen problems but exhibits a clear performance degradation on unseen variants, pushing the average gap up to 5.05\%. This performance drop occurs because FGE tends to homogenize the observation space, thereby blurring the critical distinctions among diverse constraints. Conversely, employing CARM alone effectively improves model performance by inherently boosting constraint awareness, reducing the average gap to 4.50\% even under the original PRE. Finally, combining both components yields the best overall performance among all settings. The integration creates a strong synergy: CARM leverages its robust constraint awareness to effectively compensate for the homogenization induced by FGE, pushing the overall average optimality gap down to 4.37\% and capturing the best solutions in 12 of the 16 evaluated scenarios. 

\begin{table}[htbp]
  \centering
  \caption{Ablation study of the FGE and CARM modules based on MTPOMO across 16 VRP variants.}
  \resizebox{0.99\textwidth}{!}{
    \begin{tabular}{cc|cccccc}
    \toprule[0.5mm]
    $\hat{\mathbf{h}}_{(C)}^{t}$   & CARM  & CVRP  & OCVRP & CVRPB & CVRPTW & CVRPL & OCVRPTW \\
    \midrule
    PRE & $\times$ & 1.85\% & 3.46\% & 1.67\% & 5.31\% & 0.48\% & 4.41\% \\
    PRE & $\checkmark$ & 1.64\% & 2.99\% & 1.25\% & 4.89\% & 0.26\% & 3.73\% \\
    FGE & $\times$ & 1.83\% & 3.51\% & 1.48\% & 4.98\% & 0.46\% & 3.93\% \\
    FGE & \checkmark & \greybg{1.59\%} & \greybg{2.87\%} & \greybg{1.11\%} & \greybg{4.54\%} & \greybg{0.20\%} & \greybg{3.14\%} \\
    \midrule
    $\hat{\mathbf{h}}_{(C)}^{t}$   & CARM  & OCVRPB & CVRPBL & CVRPLTW & OCVRPBTW & CVRPBLTW & OCVRPL \\
    \midrule
    PRE & $\times$ & 7.34\% & 1.79\% & 1.92\% & 10.45\% & 7.75\% & 3.44\% \\
    PRE & $\checkmark$ & 6.52\% & 1.41\% & 1.59\% & \greybg{9.93\%} & \greybg{7.36\%} & 2.94\% \\
    FGE & $\times$ & 8.06\% & 1.44\% & 1.57\% & 10.96\% & 8.17\% & 3.51\% \\
    FGE & \checkmark & \greybg{6.25\%} & \greybg{1.11\%} & \greybg{1.20\%} & 10.14\% & 7.70\% & \greybg{2.86\%} \\
    \midrule
    $\hat{\mathbf{h}}_{(C)}^{t}$   & CARM  & OCVRPBLTW & CVRPBTW & OCVRPBL & OCVRPLTW & Avg.  & Best Sol. \\
    \midrule
    PRE & $\times$ & 10.50\% & 7.41\% & 7.34\% & 4.37\% & 4.97\% & 0/16 \\
    PRE & $\checkmark$ & \greybg{9.95\%} & \greybg{7.16\%} & 6.51\% & 3.82\% & 4.50\% & 4/16 \\
    FGE & $\times$ & 11.03\% & 7.83\% & 8.11\% & 3.97\% & 5.05\% & 0/16 \\
    FGE & \checkmark & 10.27\% & 7.39\% & \greybg{6.36\%} & \greybg{3.22\%} & \greybg{4.37\%} & \greybg{12/16} \\
    \bottomrule[0.5mm]
    \end{tabular}%
    }
  \label{tab:ablation_mtpomo}%
\end{table}%

\subsection{Ablation Study on Structural Design versus Parameter Scale}
To verify that the performance gain of CARM stems from its structural design rather than a mere increase in parameter count, we conduct two additional experiments based on the MTPOMO\citep{liu2024mtpomo}. Specifically, we introduce two linear substitution variants: PreLinear, which expands the parameter size of the context embedding, and PostLinear, which increases the parameters of the subsequent state embedding. As detailed in Table \ref{tab:params_ablation}, both linear variants introduce significantly more additional parameters ($\Delta$ Params = 33.02k) compared to the CARM module ($\Delta$ Params = 17.79k). However, neither of these parameter-heavy variants matches the performance of CARM. CARM consistently outperforms all other variants, achieving the lowest average optimality gap of 4.37\% and securing the best solutions in 15 out of the 16 evaluated variants. This strongly demonstrates that the superiority of CARM is attributed to its effective constraint-aware mechanism rather than a naive expansion of the parameter space.

\begin{table}[htbp]
  \centering
  \caption{Ablation study validating the structural advantage of CARM over linear substitutions with larger parameter counts.}
  \resizebox{1\textwidth}{!}{
    \begin{tabular}{l|ccccc}
    \toprule[0.5mm]
    Strategy     & CVRP  & OCVRP & CVRPB & CVRPTW & CVRPL \\
    \midrule
    FGE-only & 1.83\% & 3.51\% & 1.48\% & 4.98\% & 0.46\% \\
    FGE+PreLinear & 1.64\% & 3.09\% & 1.20\% & 4.73\% & 0.29\% \\
    FGE+PostLinear & 1.77\% & 3.39\% & 1.36\% & 4.82\% & 0.41\% \\
    FGE+CARM & \greybg{1.59\%} & \greybg{2.87\%} & \greybg{1.11\%} & \greybg{4.54\%} & \greybg{0.20\%} \\
    \midrule
          & OCVRPTW & OCVRPB & CVRPBL & CVRPLTW & OCVRPBTW \\
    \midrule
    FGE-only & 3.93\% & 8.06\% & 1.44\% & 1.57\% & 10.96\% \\
    FGE+PreLinear & 3.36\% & 6.76\% & 1.22\% & 1.39\% & 10.50\% \\
    FGE+PostLinear & 3.69\% & 7.32\% & 1.38\% & 1.50\% & 10.76\% \\
    FGE+CARM & \greybg{3.14\%} & \greybg{6.25\%} & \greybg{1.11\%} & \greybg{1.20\%} & \greybg{10.14\%} \\
    \midrule
          & CVRPBLTW & OCVRPL & OCVRPBLTW & CVRPBTW & OCVRPBL \\
    \midrule
    FGE-only & 8.17\% & 3.99\% & 11.03\% & 8.24\% & 8.11\% \\
    FGE+PreLinear & 7.77\% & 3.07\% & 10.64\% & \greybg{7.34\%} & 6.80\% \\
    FGE+PostLinear & 7.98\% & 3.36\% & 10.89\% & 7.56\% & 7.45\% \\
    FGE+CARM & \greybg{7.70\%} & \greybg{2.86\%} & \greybg{10.27\%} & 7.39\% & \greybg{6.36\%} \\
    \midrule
          & OCVRPLTW & Params. & $\Delta$ Params. & Avg.Gap & Best Sol. \\
    \midrule
    FGE-only & 3.97\% & 1.25M & 0     & 5.11\% & 0/16 \\
    FGE+PreLinear & 3.44\% & 1.28M & 33.02K & 4.58\% & 1/16 \\
    FGE+PostLinear & 3.80\% & 1.28M & 33.02K & 4.84\% & 0/16 \\
    FGE+CARM & \greybg{3.22\%} & 1.27M & 17.79K & \greybg{4.37\%} & \greybg{15/16} \\
    \bottomrule[0.5mm]
    \end{tabular}%
  }
  \label{tab:params_ablation}%
\end{table}%

\subsection{Ablation Study on Enhancing ReLD with CARM}
To further clarify the source of performance improvements when incorporating our proposed CARM into the ReLD\citep{huang2025reld}, we conduct an additional ablation study on 10 unseen VRP variants, as detailed in Table \ref{tab:ablation_reld}. Specifically, this experiment aims to verify that the generalization gains are driven by the constraint-aware mechanism of CARM rather than the inherent identity mapping (IDT) utilized in the original ReLD architecture.

The results demonstrate that removing the identity connection (ReLD-MTL-CARM w/o IDT) still yields a substantial improvement over the basic ReLD-MTL-FGE configuration, reducing the average optimality gap from 4.98\% to 4.85\% and securing the best solutions in 5 out of the 10 unseen tasks. This explicitly confirms that the primary performance boost is derived fundamentally from CARM. Furthermore, restoring the idt connection (ReLD-MTL-CARM) to strictly adhere to the minimal modification principle of the original ReLD architecture provides an additional, albeit marginal, enhancement (further lowering the average gap to 4.83\%). Together, these findings validate that while the idt connection offers a slight structural benefit, the robust generalization capability on unseen variants is intrinsically attributed to the CARM module.

\begin{table}[htbp]
  \centering
  \caption{Generalization performance of ReLD-MTL integrated with different components (FGE, CARM w/o IDT, and CARM) evaluated on 10 unseen VRP variants.}
  \resizebox{0.99\textwidth}{!}{
    \begin{tabular}{c|cccccc}
    \toprule[0.5mm]
          & OCVRPB & CVRPBL & CVRPLTW & OCVRPBTW & CVRPBLTW & OCVRPL \\
    \midrule
    ReLD-MTL-FGE & 5.31\% & 0.69\% & 0.81\% & 9.50\% & 7.04\% & 2.28\% \\
    ReLD-MTL-CARM w/o IDT & \greybg{5.05\%} & 0.65\% & \greybg{0.69\%} & 9.41\% & 6.98\% & \greybg{2.10\%} \\
    ReLD-MTL-CARM & 5.09\% & \greybg{0.65\%} & 0.71\% & \greybg{9.30\%} & \greybg{6.92\%} & 2.14\% \\
    \midrule
          & OCVRPBLTW & CVRPBTW & OCVRPBL & OCVRPLTW & Avg.  & Best Sol. \\
    \midrule
    ReLD-MTL-FGE & 9.50\% & 6.71\% & 5.31\% & 2.63\% & 4.98\% & 0/10 \\
    ReLD-MTL-CARM w/o IDT & 9.51\% & 6.56\% & \greybg{5.08\%} & \greybg{2.43\%} & 4.85\% & 5/10 \\
    ReLD-MTL-CARM & \greybg{9.35\%} & \greybg{6.52\%} & 5.13\% & 2.50\% & \greybg{4.83\%} & \greybg{5/10} \\
    \bottomrule[0.5mm]
    \end{tabular}%
    }
  \label{tab:ablation_reld}%
\end{table}%

\clearpage
\section{Detailed Comparison of Multi-Task Solvers with PRE, FGE, and CARM}
\label{app:all_results}

Due to space constraints in the main text, we present the complete experimental results in this section. Tables \ref{tab:mtpomo_PRE_FGE_CARM} through \ref{tab:cada_PRE_FGE_CARM} provide a detailed comparison of the optimality gaps for five representative multi-task solvers: MTPOMO\citep{liu2024mtpomo}, MVMoE\citep{zhou2024mvmoe}, ReLD\citep{huang2025reld}, RouteFinder\citep{berto2024routefinder} (RF), and CaDA\citep{li2025cada}, each equipped with the PRE, FGE, and CARM configurations. Note that while MTPOMO, MVMoE, and ReLD are evaluated across 16 standard VRP variants, the evaluation environments for RF and CaDA are further extended to 24 variants, incorporating 8 additional tasks featuring the MB constraint (e.g., CVRPMB, OCVRPMB, and their combinations). This variation in the number of evaluated tasks arises from the distinct definitions of the Backhaul constraint adopted by the respective baseline models. For a detailed explanation of these definitions, please refer to Appendix \ref{app:setups_of_vrp_variants}. Across all baselines and evaluated tasks, the CARM module consistently demonstrates superior generalization and constraint-aware capabilities.

\begin{table}[htbp]
  \centering
  \caption{Detailed comparison of optimality gaps for MTPOMO under PRE, FGE, and CARM configurations across 16 VRP variants.}
  \resizebox{1\textwidth}{!}{
    \begin{tabular}{l|cccccc}
    \toprule[0.5mm]
    Method & CVRP  & OCVRP & CVRPB & CVRPTW & CVRPL & OCVRPTW \\
    \midrule
    MTPOMO-PRE & 1.85\% & 3.46\% & 1.67\% & 5.31\% & 0.48\% & 4.41\% \\
    MTPOMO-FGE & 1.83\% & 3.51\% & 1.48\% & 4.98\% & 0.46\% & 3.93\% \\
    MTPOMO-CARM & \greybg{1.59\%} & \greybg{2.87\%} & \greybg{1.11\%} & \greybg{4.54\%} & \greybg{0.20\%} & \greybg{3.14\%} \\
    \midrule
          & OCVRPB & CVRPBL & CVRPLTW & OCVRPBTW & CVRPBLTW & OCVRPL \\
    \midrule
    MTPOMO-PRE & 7.34\% & 1.79\% & 1.92\% & 10.45\% & 7.75\% & 3.44\% \\
    MTPOMO-FGE & 8.06\% & 1.44\% & 1.57\% & 10.96\% & 8.17\% & 3.99\% \\
    MTPOMO-CARM & \greybg{6.25\%} & \greybg{1.11\%} & \greybg{1.20\%} & \greybg{10.14\%} & \greybg{7.70\%} & \greybg{2.86\%} \\
    \midrule
          & OCVRPBLTW & CVRPBTW & OCVRPBL & OCVRPLTW & Avg.  & Best Sol. \\
    \midrule
    MTPOMO-PRE & 10.50\% & 7.41\% & 7.34\% & 4.37\% & 4.97\% & 0/16 \\
    MTPOMO-FGE & 11.03\% & 8.24\% & 8.11\% & 3.97\% & 5.11\% & 0/16 \\
    MTPOMO-CARM & \greybg{10.27\%} & \greybg{7.39\%} & \greybg{6.36\%} & \greybg{3.22\%} & \greybg{4.37\%} & \greybg{16/16} \\
    \bottomrule[0.5mm]
    \end{tabular}%
  }
  \label{tab:mtpomo_PRE_FGE_CARM}%
\end{table}%

\begin{table}[htbp]
  \centering
  \caption{Detailed comparison of optimality gaps for MVMoE under PRE, FGE, and CARM configurations across 16 VRP variants.}
  \resizebox{1\textwidth}{!}{
    \begin{tabular}{l|cccccc}
    \toprule[0.5mm]
    Method & CVRP  & OCVRP & CVRPB & CVRPTW & CVRPL & OCVRPTW \\
    \midrule
    MVMoE-PRE & 1.73\% & 3.21\% & 1.37\% & 4.93\% & 0.32\% & 3.94\% \\
    MVMoE-FGE & 1.65\% & 3.09\% & 1.19\% & 4.59\% & 0.26\% & 3.39\% \\
    MVMoE-CARM & \greybg{1.54\%} & \greybg{2.77\%} & \greybg{1.00\%} & \greybg{4.39\%} & \greybg{0.16\%} & \greybg{2.88\%} \\
    \midrule
          & OCVRPB & CVRPBL & CVRPLTW & OCVRPBTW & CVRPBLTW & OCVRPL \\
    \midrule
    MVMoE-PRE & 7.24\% & 1.47\% & 1.55\% & 10.19\% & 7.47\% & 3.24\% \\
    MVMoE-FGE & 7.00\% & 1.20\% & 1.23\% & 10.39\% & 7.76\% & 3.13\% \\
    MVMoE-CARM & \greybg{6.19\%} & \greybg{0.98\%} & \greybg{1.00\%} & \greybg{9.69\%} & \greybg{7.29\%} & \greybg{2.69\%} \\
    \midrule
          & OCVRPBLTW & CVRPBTW & OCVRPBL & OCVRPLTW & Avg.  & Best Sol. \\
    \midrule
    MVMoE-PRE & 10.26\% & 7.19\% & 7.30\% & 3.97\% & 4.71\% & 0/16 \\
    MVMoE-FGE & 10.45\% & 7.35\% & 7.08\% & 3.48\% & 4.58\% & 0/16 \\
    MVMoE-CARM & \greybg{9.80\%} & \greybg{6.92\%} & \greybg{6.26\%} & \greybg{2.99\%} & \greybg{4.16\%} & \greybg{16/16} \\
    \bottomrule[0.5mm]
    \end{tabular}%
  }
  \label{tab:mvmoe_PRE_FGE_CARM}%
\end{table}%

\begin{table}[htbp]
  \centering
  \caption{Detailed comparison of optimality gaps for ReLD-MTL under PRE, FGE, and CARM configurations across 16 VRP variants.}
  \resizebox{1\textwidth}{!}{
    \begin{tabular}{l|cccccc}
    \toprule[0.5mm]
    Method & CVRP  & OCVRP & CVRPB & CVRPTW & CVRPL & OCVRPTW \\
    \midrule
    ReLD-MTL-PRE & 1.42\% & 2.32\% & 0.90\% & 4.56\% & 0.02\% & 3.10\% \\
    ReLD-MTL-FGE & 1.37\% & 2.26\% & 0.68\% & 4.16\% & -0.02\% & 2.62\% \\
    ReLD-MTL-CARM & \greybg{1.34\%} & \greybg{2.11\%} & \greybg{0.67\%} & \greybg{4.09\%} & \greybg{-0.04\%} & \greybg{2.49\%} \\
    \midrule
          & OCVRPB & CVRPBL & CVRPLTW & OCVRPBTW & CVRPBLTW & OCVRPL \\
    \midrule
    ReLD-MTL-PRE & 5.36\% & 1.01\% & 1.17\% & \greybg{9.29\%} & 6.94\% & 2.31\% \\
    ReLD-MTL-FGE & 5.31\% & 0.69\% & 0.81\% & 9.50\% & 7.04\% & 2.28\% \\
    ReLD-MTL-CARM & \greybg{5.09\%} & \greybg{0.65\%} & \greybg{0.71\%} & 9.30\% & \greybg{6.92\%} & \greybg{2.14\%} \\
    \midrule
          & OCVRPBLTW & CVRPBTW & OCVRPBL & OCVRPLTW & Avg.  & Best Sol. \\
    \midrule
    ReLD-MTL-PRE & \greybg{9.22\%} & 6.74\% & 5.41\% & 3.16\% & 3.93\% & 2/16 \\
    ReLD-MTL-FGE & 9.50\% & 6.71\% & 5.31\% & 2.63\% & 3.80\% & 0/16 \\
    ReLD-MTL-CARM & 9.35\% & \greybg{6.52\%} & \greybg{5.13\%} & \greybg{2.50\%} & \greybg{3.68\%} & \greybg{14/16} \\
    \bottomrule[0.5mm]
    \end{tabular}%
  }
  \label{tab:reld_PRE_FGE_CARM}%
\end{table}%

\begin{table}[htbp]
  \centering
  \caption{Detailed comparison of optimality gaps for RouteFinder under PRE, FGE, and CARM configurations across 24 VRP variants.}
  \resizebox{1\textwidth}{!}{
    \begin{tabular}{l|cccccc}
    \toprule[0.5mm]
    Method & CVRP  & OCVRP & CVRPB & CVRPTW & CVRPL & OCVRPTW \\
    \midrule
    RF-PRE & \greybg{1.51\%} & 4.06\% & 3.95\% & 3.18\% & 1.83\% & 2.35\% \\
    RF-FGE & 1.67\% & 4.16\% & 4.02\% & 2.44\% & 1.98\% & \greybg{1.60\%} \\
    RF-CARM & 1.53\% & \greybg{3.94\%} & \greybg{3.85\%} & \greybg{2.42\%} & \greybg{1.82\%} & 1.62\% \\
    \midrule
          & OCVRPB & CVRPBL & CVRPLTW & OCVRPBTW & CVRPBLTW & OCVRPL \\
    \midrule
    RF-PRE & 4.21\% & 5.04\% & 3.58\% & 2.04\% & 2.92\% & 4.05\% \\
    RF-FGE & 3.96\% & 5.00\% & 2.85\% & \greybg{1.35\%} & 2.31\% & 4.17\% \\
    RF-CARM & \greybg{3.76\%} & \greybg{4.80\%} & \greybg{2.84\%} & 1.38\% & \greybg{2.29\%} & \greybg{3.94\%} \\
    \midrule
          & OCVRPBLTW & CVRPBTW & OCVRPBL & OCVRPLTW & CVRPMB & OCVRPMB \\
    \midrule
    RF-PRE & 2.05\% & 2.62\% & 4.26\% & 2.35\% & 10.18\% & 18.90\% \\
    RF-FGE & \greybg{1.36\%} & \greybg{1.90\%} & 3.98\% & \greybg{1.60\%} & 9.83\% & 17.86\% \\
    RF-CARM & 1.39\% & 1.94\% & \greybg{3.79\%} & 1.63\% & \greybg{9.65\%} & \greybg{17.51\%} \\
    \midrule
          & CVRPMBL & OCVRPMBL & CVRPMBTW & OCVRPMBTW & CVRPMBLTW & OCVRPMBLTW \\
    \midrule
    RF-PRE & 10.37\% & 18.87\% & 10.84\% & \greybg{8.73\%} & 10.81\% & \greybg{8.73\%} \\
    RF-FGE & 10.20\% & 17.85\% & 11.12\% & 9.41\% & 11.23\% & 9.42\% \\
    RF-CARM & \greybg{9.62\%} & \greybg{17.44\%} & \greybg{10.48\%} & 8.74\% & \greybg{10.65\%} & 8.84\% \\
    \bottomrule[0.5mm]
    \end{tabular}%
  }
  \label{tab:rf_PRE_FGE_CARM}%
\end{table}%

\begin{table}[htbp]
  \centering
  \caption{Detailed comparison of optimality gaps for CaDA under PRE, FGE, and CARM configurations across 24 VRP variants.}
  \resizebox{1\textwidth}{!}{
    \begin{tabular}{l|cccccc}
    \toprule[0.5mm]
    Method & CVRP  & OCVRP & CVRPB & CVRPTW & CVRPL & OCVRPTW \\
    \midrule
    CaDA-PRE & 1.52\% & 4.00\% & 3.91\% & 3.15\% & 1.86\% & 2.35\% \\
    CaDA-FGE & 1.58\% & 4.14\% & 4.01\% & 2.18\% & 1.86\% & 1.43\% \\
    CaDA-CARM & \greybg{1.50\%} & \greybg{3.98\%} & \greybg{3.74\%} & \greybg{2.11\%} & \greybg{1.76\%} & \greybg{1.35\%} \\
    \midrule
          & OCVRPB & CVRPBL & CVRPLTW & OCVRPBTW & CVRPBLTW & OCVRPL \\
    \midrule
    CaDA-PRE & 4.10\% & 5.00\% & 3.00\% & 1.44\% & 2.36\% & 4.05\% \\
    CaDA-FGE & 3.89\% & 4.99\% & 2.56\% & 1.18\% & 2.07\% & 4.12\% \\
    CaDA-CARM & \greybg{3.79\%} & \greybg{4.73\%} & \greybg{2.46\%} & \greybg{1.11\%} & \greybg{1.98\%} & \greybg{3.97\%} \\
    \midrule
          & OCVRPBLTW & CVRPBTW & OCVRPBL & OCVRPLTW & CVRPMB & OCVRPMB \\
    \midrule
    CaDA-PRE & 1.44\% & 2.00\% & 4.10\% & 1.75\% & 9.69\% & 17.38\% \\
    CaDA-FGE & 1.19\% & 1.71\% & 3.93\% & 1.42\% & 9.46\% & \greybg{16.47\%} \\
    CaDA-CARM & \greybg{1.11\%} & \greybg{1.63\%} & \greybg{3.82\%} & \greybg{1.35\%} & \greybg{9.34\%} & 16.66\% \\
    \midrule
          & CVRPMBL & OCVRPMBL & CVRPMBTW & OCVRPMBTW & CVRPMBLTW & OCVRPMBLTW \\
    \midrule
    CaDA-PRE & \greybg{9.29\%} & 17.39\% & 11.53\% & 9.49\% & 11.45\% & 9.48\% \\
    CaDA-FGE & 9.76\% & \greybg{16.47\%} & 10.73\% & 9.09\% & 10.86\% & 9.08\% \\
    CaDA-CARM & 9.40\% & 16.68\% & \greybg{10.30\%} & \greybg{8.89\%} & \greybg{10.43\%} & \greybg{8.90\%} \\
    \bottomrule[0.5mm]
    \end{tabular}%
  }
  \label{tab:cada_PRE_FGE_CARM}%
\end{table}%

\clearpage
\section{Licenses For Used Resources}
\label{append:licenses}

\begin{table}[htbp]
\centering
\caption{List of licenses for the codes and datasets we used in this work.}
\label{table:Licenses}
\resizebox{0.99\textwidth}{!}{%
\begin{tabular}{l |l | l | l }
\toprule[0.5mm]
 Resource   &   Type  &  Link  & License    \\
\midrule
HGS-PyVRP \citep{wouda2024pyvrp}& Code & \url{https://github.com/PyVRP/PyVRP} & MIT License\\
LKH3 \citep{LKH3} & Code & \url{http://webhotel4.ruc.dk/~keld/research/LKH-3/} & Available for academic research use\\

OR-Tools \citep{ortools}& Code & \url{https://github.com/google/or-tools} & Apache-2.0 License\\
\midrule

MTPOMO \citep{liu2024mtpomo} & Code & \url{https://github.com/FeiLiu36/MTNCO} & MIT License  \\
MVMoE \citep{zhou2024mvmoe} & Code & \url{https://github.com/RoyalSkye/Routing-MVMoE} & MIT License  \\
RouteFinder \citep{berto2024routefinder} & Code & \url{https://github.com/ai4co/routefinder} & MIT License  \\
CaDA \citep{li2025cada} & Code & \url{https://github.com/CIAM-Group/CaDA} & MIT License  \\
ReLD \citep{huang2025reld} & Code & \url{https://github.com/ziweileonhuang/reld-nco} & MIT License  \\
\midrule
Large-Scale Multi-Variant VRP Dataset \citep{zheng2025mtl_kd} & Dataset & \url{https://drive.google.com/drive/folders/1E_pu4a7BWdRPhbGcLJkQ86-74B-X2ibh?usp=sharing} & MIT License  \\
Multi-task VRP dataset (MVMoE Version) \citep{zhou2024mvmoe} & Dataset & \url{https://github.com/RoyalSkye/Routing-MVMoE/tree/main/data} & MIT License  \\
Multi-task VRP dataset (RouteFinder Version) \citep{berto2024routefinder} & Dataset & \url{https://huggingface.co/datasets/ai4co/routefinder} & MIT License  \\
\bottomrule[0.5mm]
\end{tabular}%
}
\end{table}

We list the used existing codes and datasets in \cref{table:Licenses}, and all of them are open-sourced resources for academic usage.

\section{Broader Impacts}
\label{app:broader_impacts}
This research contributes to the field of neural combinatorial optimization by employing advanced machine learning techniques to address diverse and large-scale VRPs. We believe that our findings regarding the preservation of a global observation space, alongside the proposed CARM module, will offer valuable insights and inspire future exploration of more efficient neural methods for unseen and multi-variant VRPs. Furthermore, as a general learning-based approach to the VRP, the CARM module does not inherently pose any specific negative social impacts.


\end{CJK*}
\end{document}